\documentclass[review,times,10pt]{elsarticle}

\usepackage{graphicx}%
\usepackage{multirow}%
\usepackage{amsmath,amssymb,amsfonts}%
\usepackage{amsthm}%
\usepackage{mathrsfs}%
\usepackage{xcolor}%
\usepackage{textcomp}%
\usepackage{manyfoot}%
\usepackage{booktabs}%
\usepackage{algorithm}%
\usepackage{algorithmicx}%
\usepackage{algpseudocode}%
\usepackage{listings}%
\usepackage{subfigure}
\usepackage{pifont}
\usepackage{bm}
\newcommand{\cmark}{\text{\ding{51}}}
\newcommand{\xmark}{\text{\ding{55}}}

\journal{elsarticle}

\begin{document}

\begin{frontmatter}


\title{Region-Guided Attack on the Segment Anything Model}




\author[1]{Xiaoliang Liu\corref{cor1}}
\ead{20249197@wzbc.edu.cn}
\cortext[cor1]{Corresponding author: Xiaoliang Liu.}

\author[2,3]{Furao Shen}
\ead{frshen@nju.edu.cn}
\author[4]{Jian Zhao}
\ead{jianzhao@nju.edu.cn}
\address[1]{School of Information Engineering, Wenzhou Business College, China}
\address[2]{National Key Laboratory for Novel Software Technology, Nanjing University, China}
\address[3]{School of Artificial Intelligence, Nanjing University, China}
\address[4]{School of Electronic Science and Engineering, Nanjing University, China}

\begin{abstract}
The Segment Anything Model (SAM) is a cornerstone of image segmentation, demonstrating exceptional performance across various applications, particularly in autonomous driving and medical imaging, where precise segmentation is crucial. However, SAM is vulnerable to adversarial attacks that can significantly impair its functionality through minor input perturbations. Traditional techniques, such as FGSM and PGD, are often ineffective in segmentation tasks due to their reliance on global perturbations that overlook spatial nuances. Recent methods like Attack-SAM-K and UAD have begun to address these challenges, but they frequently depend on external cues and do not fully leverage the structural interdependencies within segmentation processes. This limitation underscores the need for a novel adversarial strategy that exploits the unique characteristics of segmentation tasks. In response, we introduce the Region-Guided Attack (RGA), designed specifically for SAM. RGA utilizes a Region-Guided Map (RGM) to manipulate segmented regions, enabling targeted perturbations that fragment large segments and expand smaller ones, resulting in erroneous outputs from SAM. Our experiments demonstrate that RGA achieves high success rates in both white-box and black-box scenarios, emphasizing the need for robust defenses against such sophisticated attacks. RGA not only reveals SAM's vulnerabilities but also lays the groundwork for developing more resilient defenses against adversarial threats in image segmentation.
\end{abstract}






\end{frontmatter}


\section{Introduction}
\label{sec:introduction}
The Segment Anything Model (SAM)~\cite{kirillov2023segment} has emerged as a leading solution in image segmentation, demonstrating remarkable adaptability and performance across diverse datasets and prompts. Its architecture allows for seamless integration with various inputs, making it a pivotal tool for applications ranging from autonomous driving~\cite{yan2024segment,ieee2024enhancing} to medical imaging~\cite{MAZUROWSKI2023102918,zhang2023customized}. However, this versatility also exposes SAM to vulnerabilities, particularly from adversarial attacks that can significantly degrade its performance~\cite{zhang2023ASK,lu2024UAD}. These attacks leverage subtle perturbations in the input data, misleading the model into producing incorrect segmentations, thereby raising concerns about the reliability of SAM in critical contexts.

Previous adversarial attack methods, such as the Fast Gradient Sign Method (FGSM)~\cite{goodfellow2015explaining} and Projected Gradient Descent (PGD)~\cite{madry2018towards}, primarily focus on classification models and often utilize global perturbations that affect the entire input. These methods can be less effective in segmentation tasks, where spatial relationships and contextual information are critical. Recent approaches like Attack-SAM-K~\cite{zhang2023ASK} and UAD~\cite{lu2024UAD} have begun to explore the unique challenges associated with adversarial attacks on segmentation models, but many still rely on external prompts or do not fully exploit the structural dependencies inherent in the segmentation process.

To effectively address these vulnerabilities, we propose the Region-Guided Attack (RGA), a novel adversarial attack strategy specifically designed for SAM. Unlike traditional adversarial methods that often rely on external prompts or global perturbations, RGA focuses on manipulating segmented regions directly through a Region-Guided Map (RGM). This approach allows for targeted adversarial perturbations that divide large segments into smaller fragments while merging smaller regions into larger areas, ultimately leading to misclassifications in SAM’s outputs. The innovation of RGA lies in its ability to exploit the structural dependencies within the segmentation task, leveraging the inherent characteristics of SAM to enhance the effectiveness of the attack.

The significance of RGA is twofold. First, it provides a deeper understanding of the vulnerabilities inherent in advanced segmentation models like SAM, offering insights into how adversarial perturbations can be crafted more strategically. Second, RGA presents a more refined method of inducing segmentation errors that can be applied across various segmentation frameworks, highlighting the need for robust defenses against such targeted attacks. Through extensive experiments, we demonstrate the effectiveness of RGA, revealing its capability to achieve high attack success rates in both white-box and black-box scenarios while maintaining minimal perceptual distortion in the input images. In summary, the key contributions of RGA include:
\begin{enumerate}
\item \textbf{Region-Guided Map (RGM) for Adversarial Guidance}: RGA introduces a novel use of the RGM to directly guide the generation of adversarial examples. By utilizing RGM to define how SAM's segmentation should be altered (i.e., splitting large regions into smaller ones and merging smaller regions into larger ones), RGA effectively guides perturbations to maximize the impact on segmentation quality.

\item \textbf{Enhanced Attack Success and Transferability}: By leveraging RGM, RGA achieves higher attack success rates and improved transferability. The adversarial examples are generated with a clear objective influenced by the segmentation output, which systematically guides perturbations, leading to more successful and transferable attacks against the SAM model.

\item \textbf{Independent of External Prompts}: Unlike many existing methods that rely heavily on specific prompts to guide attacks, RGA operates independently of external prompts, making the adversarial process more streamlined and broadly applicable. This independence ensures that RGA can be applied in scenarios where prompts are unavailable or unpredictable.

\item \textbf{Insights into SAM’s Segmentation Vulnerabilities}: RGA reveals particular vulnerabilities in SAM by focusing on regional manipulations instead of global input perturbations. The findings highlight how region-specific guidance, such as altering the size and boundaries of segmented areas, can degrade SAM's segmentation performance significantly. This understanding provides valuable insights for designing more resilient segmentation models.
\end{enumerate}

\section{Related Works}
\label{sec:related_works}
\subsection{Segmentation Models}
Image segmentation is a critical task in computer vision, aiming to partition an image into meaningful segments or objects at the pixel level. Traditional approaches relied heavily on hand-crafted features and were limited in handling complex scenes. The advent of deep learning revolutionized segmentation tasks with models like Fully Convolutional Networks (FCNs)~\cite{long2015fully}, which replaced fully connected layers with convolutional ones to maintain spatial information.

Building upon FCNs, the U-Net architecture~\cite{ronneberger2015u} introduced an encoder-decoder structure with skip connections, enabling precise localization and context assimilation, particularly in biomedical image segmentation. DeepLab models~\cite{chen2018deeplab} further enhanced segmentation by incorporating atrous convolution and conditional random fields for capturing multi-scale context.

The SAM, introduced by Meta AI in 2023~\cite{kirillov2023segment}, represents a significant leap in segmentation models. SAM is designed as a promptable segmentation system that can generate high-quality object masks from user input prompts, such as points, boxes, or text descriptions. Trained on a massive dataset of over one billion masks, SAM demonstrates remarkable generalization across diverse image distributions and tasks without the need for additional training.

SAM's architecture comprises three main components: an image encoder, a flexible prompt encoder, and a fast mask decoder. The image encoder processes the input image to produce an embedding, the prompt encoder transforms user prompts into embeddings, and the mask decoder combines these embeddings to generate segmentation masks. This design allows SAM to perform zero-shot generalization to new tasks and image domains, making it a foundational model for various segmentation applications.

Subsequent research has focused on adapting SAM to specific domains and improving its efficiency. For instance, efforts have been made to fine-tune SAM for medical image segmentation, where domain-specific features are crucial~\cite{ma2024segment}. Other studies explore integrating SAM with text-based prompts to enhance interactive segmentation capabilities~\cite{cheng2021per}.

\subsection{Adversarial Attacks}
Adversarial attacks deliberately manipulate input data to deceive machine learning models into making incorrect predictions. Initially studied in image classification~\cite{szegedy2013intriguing, goodfellow2015explaining}, these attacks exploit the vulnerability of deep neural networks to small, imperceptible perturbations.

The Fast Gradient Sign Method (FGSM)~\cite{goodfellow2015explaining} was one of the first techniques introduced to generate adversarial examples by performing a one-step gradient update along the direction of the gradient's sign of the loss function with respect to the input image. Its iterative variant, the Basic Iterative Method (BIM) or I-FGSM~\cite{kurakin2017adversarial}, applies FGSM multiple times with smaller step sizes, increasing the attack's success rate.

The Projected Gradient Descent (PGD)~\cite{madry2018towards} attack extends BIM by adding random initialization within the allowed perturbation radius and projecting the adversarial example back onto the feasible set after each iteration. PGD is considered a universal "first-order adversary" and is widely used due to its effectiveness in finding robust adversarial examples.

To enhance the effectiveness and transferability of adversarial examples, several advanced methods build upon FGSM, BIM, and PGD:

Momentum Iterative Fast Gradient Sign Method (MIM)~\cite{dong2018boosting}: Incorporates a momentum term into the iterative attack process, stabilizing update directions and improving transferability against different models.

Diverse Input Iterative Fast Gradient Sign Method (DIM)~\cite{xie2019improving}: DIM increases the diversity of adversarial examples by applying random transformations, such as image resizing and padding, to the input image at each iteration. This randomization helps adversarial perturbations remain effective across models that process inputs of varying dimensions or padding schemes.

Translation-Invariant Iterative Fast Gradient Sign Method (TIM)~\cite{dong2019evading}: Crafts perturbations invariant to image translations by convolving the gradient with a predefined kernel, increasing transferability across models with different architectures and training data.

Scale-Invariant Iterative Fast Gradient Sign Method (SIM)~\cite{lin2019nesterov}: Averages gradients over multiple scaled copies of the input image, capturing scale variations and making adversarial examples effective against models processing images at various scales.

RSTAM~\cite{liu2022rstam}, known as Random Similarity Transformation (RST) Attack Method, applies random similarity transformations (including translation, rotation, and scaling) to diversify the input image during the adversarial attack. This method increases the transferability of adversarial examples by ensuring that perturbations remain effective even under a variety of geometric transformations. EAP~\cite{liu2024eap} further innovated by incorporating an image pyramid and meta-ensemble strategy into the RST framework.

However, these methods have not been specifically designed to attack the SAM. Given SAM's unique architecture and prompt-based segmentation capabilities, there is a need to explore adversarial attacks tailored to SAM. In this work, we aim to investigate and develop adversarial attack strategies specifically targeting SAM, to better understand its vulnerabilities and improve its robustness.

\subsection{Adversarial Attacks on Segmentation Models}
Recent advancements in adversarial attacks on segmentation models have introduced several methods that challenge model robustness and expose vulnerabilities in complex feature extraction processes. Attack-SAM-K (ASK)~\cite{zhang2023ASK} employs a global reduction of feature responses by utilizing  $K$ point prompts distributed across the entire image, with $K$ often set to a large value such as 400. This approach is designed to manipulate the SAM model’s segmentation responses on a broad scale, directly targeting its feature extraction pipeline.

Transferable Adversarial Perturbations (TAP)~\cite{zhou2018TAP} introduces perturbations that push adversarial features away from original features using Minkowski distance. By focusing on perturbation transferability, TAP highlights cross-model vulnerabilities, making it effective across various model architectures. Building upon TAP, Intermediate-Level Perturbation Decay (ILPD)~\cite{li2023ILPD} refines this approach by maintaining an effective adversarial direction with an increased perturbation magnitude. ILPD targets intermediate-level features, thereby testing the resilience of models at deeper feature layers.

Another method, Activation Attack (AA)~\cite{Inkawhich2019AA}, minimizes the distance between adversarial features and target image features. By manipulating the activation layers, AA achieves targeted feature alignment, granting precise control over model outputs. Prompt-Agnostic Target Attack (PATA)~\cite{zheng2023PATA} extends AA by incorporating a regularization term to boost the feature dominance of adversarial images over randomly selected clean images. This modification enhances the flexibility of the attack, making it prompt-agnostic and applicable across diverse input conditions.

An extension of PATA, PATA++~\cite{zheng2023PATA}, addresses the inherent conflict between feature similarity and dominance. It achieves this by selecting a new competition image during each adversarial update iteration, which dynamically adapts the attack to optimize adversarial effectiveness through iteration-based adjustments.

Lastly, Unsegment Anything by Simulating Deformation (UAD)~\cite{lu2024UAD} is a technique that focuses on disrupting the SAM model's segmentation by simulating structural deformations within the image. UAD alters the image's structural details while maintaining an effective adversarial feature distance, using an optimized differentiable deformation function. This approach enhances the robustness and transferability of adversarial examples, offering an effective means of challenging segmentation models across varying structural conditions.

Despite these advancements, the transferability of adversarial attacks specifically designed for SAM still requires improvement. Previous methods may not generalize well across different models or real-world scenarios. In this work, we aim to address this limitation by developing a novel attack strategy that enhances transferability against SAM.

\begin{figure*}
    \centering
    \includegraphics[width=\linewidth]{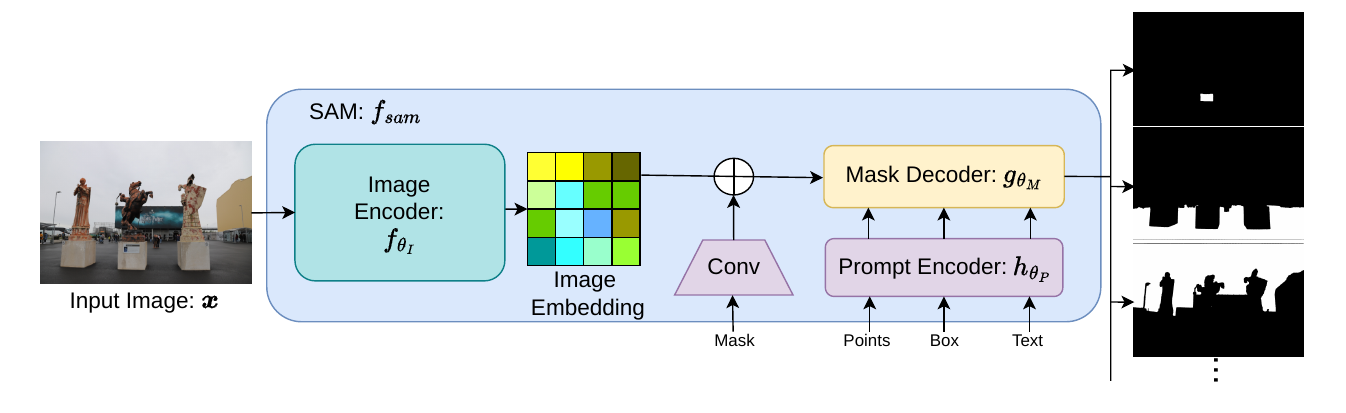}
    \caption{Overview of the architecture of the Segment Anything Model (SAM), highlighting its key components and operational flow.}
    \label{fig:sam}
\end{figure*}

\section{Preliminary}\label{sec3}
In this section, we establish the foundational concepts and framework for understanding the SAM and its susceptibility to adversarial attacks. 

\subsection{Architecture of the Segment Anything Model}
The SAM is a promptable segmentation model designed to handle a wide range of image segmentation tasks.  Its architecture comprises three primary components, as illustrated in the Figure~\ref{fig:sam}:

Image Encoder $f_{\theta_{I}}$: The image encoder processes the input image $\boldsymbol{x}$ and extracts high-dimensional feature representations. These feature embeddings encapsulate essential visual details required for accurate segmentation.

Prompt Encoder $h_{\theta_P}$: SAM accepts various forms of prompts, such as points, bounding boxes, and textual descriptions. The prompt encoder transforms these user-defined prompts $P$ into embeddings that guide the segmentation process, enabling SAM to tackle diverse tasks without necessitating retraining.

Mask Decoder $g_{\theta_M}$: The mask decoder integrates the outputs from the image encoder and prompt encoder to generate the final segmentation mask. This lightweight component ensures efficient and rapid mask generation.

SAM's ability to adapt to different types of inputs and tasks makes it a powerful tool, but it also exposes the model to adversarial vulnerabilities. The model's reliance on both image and prompt encoders to generate masks means that small perturbations to the input image or the prompt embeddings can lead to significant segmentation errors.

\subsection{Adversarial Attacks in Segmentation}

Adversarial attacks in segmentation involve generating small, imperceptible perturbations to the input image that mislead the model into producing incorrect segmentation results. These attacks, originally studied in classification tasks, have been extended to segmentation models. Common methods like FGSM~\cite{goodfellow2015explaining} and PGD~\cite{madry2018towards} target pixel-level predictions, making them applicable in segmentation tasks.

For SAM, adversarial attacks must account for how the model processes both the image and prompt embeddings. The challenge is to craft perturbations that disrupt SAM’s ability to correctly combine these embeddings, leading to erroneous segmentation masks. The goal is to minimize the model's Intersection over Union (IoU) score, effectively degrading its segmentation performance.

The perturbations are typically constrained by an $\ell_\infty$-norm, which limits their magnitude to remain imperceptible to human observers while causing significant model errors. Formally, the adversarial objective can be defined as:

\begin{equation}
\delta = \mathop{\text{argmix}}_{\delta, \|\delta\|_\infty \leq \epsilon} \mathbb{E}_{\theta_{I}, \theta_P, \theta_M}\left[\text{IoU}(g_{\theta_M}(f_{\theta_{\boldsymbol{x}}}(\boldsymbol{x} + \delta), h_{\theta_P}(P)), g_{\theta_M}(f_{\theta_{\boldsymbol{x}}}(\boldsymbol{x}), h_{\theta_P}(P)))\right] 
\end{equation}
where $\delta$ is the adversarial perturbation, $\boldsymbol{x}$ is the input image, $P$ is the prompt, and $\epsilon$ is the perturbation limit. The function $g_{\theta_M}$ represents the mask decoder, $f_{\theta_{I}}$ the image encoder, and $h_{\theta_P}$ the prompt encoder. 

The adversary aims to find a perturbation $\delta$ that minimizes the IoU between the original mask and the mask resulting from the perturbed image, effectively disrupting SAM's output while keeping the perturbation imperceptible.

\subsection{Challenges in Adversarial Attacks on SAM}

Adversarial attacks on SAM present unique challenges compared to standard segmentation models:

\begin{itemize}
    \item \textbf{Prompt-Agnostic Attacks}: SAM’s versatility in handling different types of prompts makes it challenging to design adversarial examples that generalize across various prompts. Unlike traditional segmentation models, where the input is static, SAM’s output depends on the specific prompts provided by the user, increasing the complexity of the attack.
    
    \item \textbf{Transferability}: Attacks designed for SAM must also transfer effectively across different segmentation models, including those with varying architectures and prompt types. Many existing attacks fail to generalize across different models, limiting their practical impact.
    
    \item \textbf{Feature-Level Perturbations}: Attacks on SAM must focus on disrupting the model's feature representations. Since SAM relies on both image and prompt features to generate masks, the adversary must craft perturbations that target the image feature space without being overly dependent on a specific prompt.
\end{itemize}

These challenges motivate the need for novel adversarial strategies, such as the Region-Guided Attack, which is designed to exploit SAM's segmentation vulnerabilities more effectively.

\begin{figure*}
    \centering
    \includegraphics[width=\linewidth]{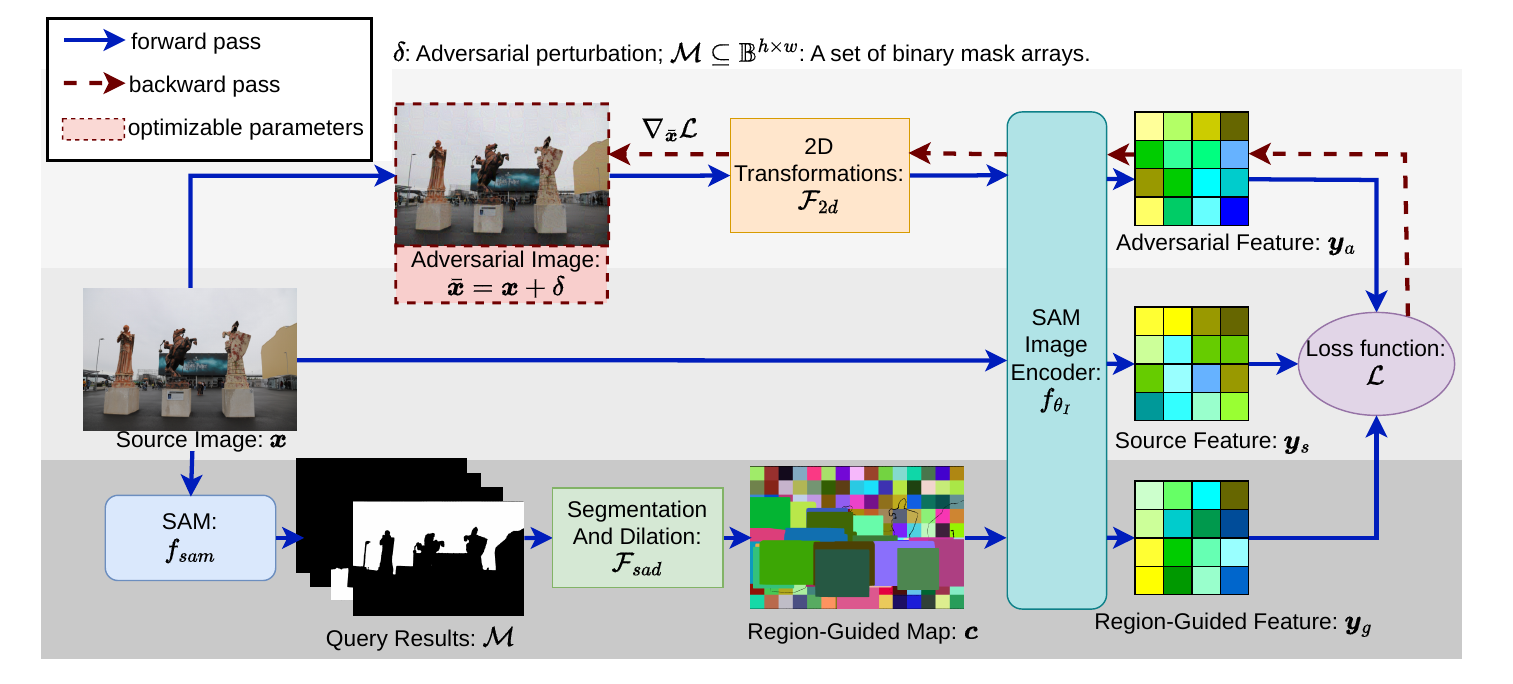}
    \caption{Overview of the Region-Guided Attack (RGA) framework on SAM. }
    \label{fig:overview}
\end{figure*}

\section{Approach}\label{sec4}
In this section, we present the details of our proposed Region-Guided Attack (RGA) targeting the Segment Anything Model (SAM). The method is designed to manipulate SAM's segmentation capabilities by systematically altering how the model interprets regions within an image. By focusing on both large and small segmented areas, we can induce errors in SAM’s output with minimal and imperceptible perturbations to the input image. The RGA framework leverages SAM’s prompt-based segmentation architecture to achieve targeted adversarial attacks while maintaining a high degree of transferability to other segmentation models.

\subsection{Overview}
The Region-Guided Attack (RGA) targets the SAM by manipulating how it segments image regions, causing segmentation errors through strategic adversarial perturbations. The attack process involves querying SAM for an initial segmentation, generating a guidance map based on the model’s output, and using the Segmentation and Dilation Strategy (SAD) to craft perturbations that alter SAM’s segmentation boundaries. This manipulation divides large regions into smaller segments and merges small regions into larger ones, disrupting the segmentation process.

As illustrated in the Figure~\ref{fig:overview}, the RGA framework is structured around three key steps:

\begin{itemize}
 \item Single Query: The process begins by feeding the original image into SAM and retrieving the segmentation result. This result provides the basis for generating the adversarial perturbations.

 \item Region-Guided Map (RGM) : RGM is constructed based on the initial segmentation result from SAM. It serves as a guide for how adversarial perturbations should be applied to influence SAM's output. Specifically, the RGM is used to misguide SAM by splitting originally segmented large regions into smaller fragments, while smaller regions are expanded into larger ones. This is achieved by leveraging the SAD strategy, which tailors the perturbation process to different region sizes — applying grid-based segmentation for large areas and dilation-based enhancement for smaller areas. The RGM effectively guides SAM to make segmentation errors, altering the size and boundaries of regions in a manner that leads to incorrect segmentation outputs.

 \item Adversarial Perturbation: The adversarial perturbations are generated through a gradient-based optimization process. The loss function calculates the difference between SAM's initial segmentation and the desired incorrect segmentation based on the guidance map. By applying these perturbations to the input image, the model is forced to produce incorrect segmentations without visually altering the input image.

\end{itemize}

The workflow involves a forward pass, where SAM generates the segmentation based on the perturbed image, and a backward pass to optimize the perturbations using the guidance map. The adversarial perturbations are then applied to the image, resulting in an adversarial image that misleads SAM, as shown in the figure~\ref{fig:overview}.

\subsection{Problem Formulation}
The goal of the RGA is to generate adversarial perturbations for an input image $\boldsymbol{x}$ such that the Segment Anything Model (SAM), denoted as $f_{\text{sam}}$, produces incorrect segmentation results that align with a predefined guidance map. Given an input image $\boldsymbol{x}$ and SAM $f_{\text{sam}}$, we want to generate a perturbation $\delta$ such that when applied to the image $\boldsymbol{x}$, the perturbed image $\boldsymbol{x} + \delta$ leads SAM to produce a segmentation output that deviates significantly from the original segmentation output, as defined by the guidance map $\boldsymbol{c}$. The aim is to guide SAM into splitting large segments into smaller fragments and merging smaller regions into larger ones, ultimately compromising segmentation accuracy. The optimization objective can be described as follows:
\begin{equation}
\delta = \mathop{\text{argmix}}_{\delta, \|\delta\|_{\infty} \leq \epsilon} \mathcal{L}\left(f_{\theta_{I}}(\boldsymbol{x}), f_{\theta_{I}}(\mathcal{F}_{\text{2d}}(\boldsymbol{x} + \delta)), f_{\theta_{I}}(\boldsymbol{c})\right)
\end{equation}
where $\delta$ represents the adversarial perturbation applied to the image $\boldsymbol{x}$, $\epsilon$ denotes the perturbation bound that ensures the perturbation remains imperceptible to human observers, $f_{\theta_{I}}$ is the image encoder function used in SAM to generate feature embeddings, $\mathcal{F}_{\text{2d}}(\cdot)$ refers to the 2D transformation function that augments the perturbed image to increase its diversity, and $\boldsymbol{c}$ is the guidance map generated using the SAD strategy, which serves as a target for directing the segmentation output towards incorrect boundaries.

The goal is to find an adversarial perturbation $\delta$ that maximizes the loss function $\mathcal{L}$, which measures the divergence between SAM's original segmentation and the target segmentation defined by $\boldsymbol{c}$. This ensures that the segmentation output is altered according to the desired adversarial effect.

\subsection{Segmentation and Dilation Strategy}

In this work, we introduce the Segmentation and Dilation (SAD) strategy for efficiently processing binary mask arrays and applying random colorization based on the size of the segmented regions. This strategy addresses varying scales of segmented areas through two distinct approaches: \textbf{grid-based segmentation} for large regions and \textbf{dilation-based enhancement} for smaller regions, ensuring accurate and non-overlapping color application across the mask.

The grid size, denoted as $grid\_size$, is calculated based on the image dimensions (width $w$ and height $h$) and the granularity parameter $\gamma$:
\begin{equation}
grid\_size = \left\lfloor \min(w, h) \times \gamma \right\rfloor
\end{equation}
where $w$ and $h$ represent the width and height of the image, respectively, and $\gamma$ controls the size of the grid blocks used for large-region segmentation.

For \textbf{small regions}, defined as those with an area smaller than $grid\_size^2$, a dilation-based enhancement operation is applied. This operation, called $\text{DilateRegion}$, is performed using a predefined structuring element over $n$ iterations, where $n$ is the number of dilation steps. A random RGB color $R_1$ is assigned to each small region, and the dilated mask is updated by applying the color only to previously uncolored regions:
\begin{equation}
\boldsymbol{c} \leftarrow \boldsymbol{c} + (R_1 \times \text{DilateRegion}(\boldsymbol{M}_j, n)) \cdot (\boldsymbol{c} == 0).
\end{equation}
Here, $\boldsymbol{c} \in \mathbb{R}^{h \times w \times 3}$ denotes the color mask, which is initialized to zeros, meaning that no areas are colored initially. $\text{DilateRegion}(\boldsymbol{M}_j, n)$ denotes the mask $\boldsymbol{M}_j$ after $n$ iterations of dilation, and $\boldsymbol{c} == 0$ ensures that the random color $R_1$ is applied solely to uncolored areas in the final mask $\boldsymbol{c}$.

For \textbf{large regions}, where the segmented area exceeds $grid\_size^2$, the region is subdivided into grid blocks of size $grid\_size$. Each grid block containing at least one positive pixel is assigned a unique random RGB color $R_2$. The color is applied to each grid block, ensuring no overlap, and the colored blocks are combined to form the final result:
\begin{equation}
\boldsymbol{c} \leftarrow \boldsymbol{c} + (R_2 \times \text{SegmentGrid}(\boldsymbol{M}_j, grid\_size)) \cdot (\boldsymbol{c} == 0),
\end{equation}
where $\text{SegmentGrid}(\boldsymbol{M}_j, grid\_size)$ refers to the grid-based segmentation of the large-region mask $\boldsymbol{M}_j$.

\begin{figure*}
    \centering
    \includegraphics[width=\linewidth]{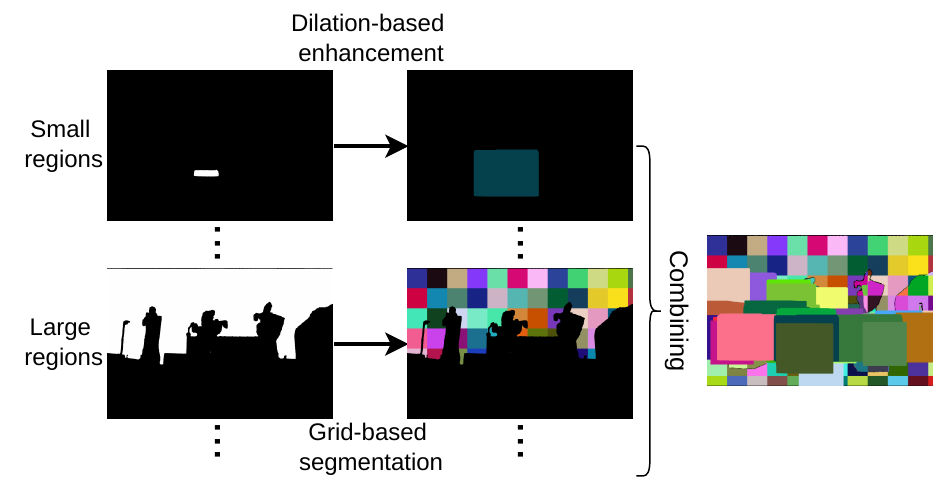}
    \caption{Example of using SAD strategy to generate region-guided map.}
    \label{fig:sad}
\end{figure*}

By combining these two techniques—dilation-based enhancement with $n$ iterations for small regions and grid-based segmentation for large regions—the SAD strategy effectively adapts to regions of various sizes. This dual approach ensures that the final mask $\boldsymbol{c}$ is accurately colorized without overlap, while enhancing the visibility of smaller regions that might otherwise be neglected.As shown in Figure~\ref{fig:sad}, this is an example of using the SAD strategy to generate a region-guided map. For more detailed descriptions, please refer to Algorithm~\ref{alg:01}.

\begin{algorithm}[!ht]
\caption{Segmentation and Dilation Strategy}
\begin{algorithmic}[1]
\Require {\begin{tabular}[t]{@{}l@{}}
$\mathcal{M} \subseteq \mathbb{B}^{h \times w}$: A set of binary mask arrays \\
$w$, $h$: Width and height of the image \\
$\gamma$: Granularity parameter for grid size \\
$n$: Number of dilation iterations
\end{tabular}}
\Ensure $\boldsymbol{c}$: Colored mask

\State Initialize color mask $\boldsymbol{c} \leftarrow \text{zeros}(h, w, 3)$ \Comment{All zeros}
\State Compute grid size $grid\_size \leftarrow \lfloor \min(w, h) \times \gamma \rfloor$

\For{each mask $\boldsymbol{M}_j$ in $\mathcal{M}$}
    \If{area of $\boldsymbol{M}_j \leq grid\_size^2$} \Comment{Small region detected}
        \State Perform dilation on small region: $\text{DilateRegion}(\boldsymbol{M}_j, n)$
        \State Generate random color $R_1$
        \State Apply color $\boldsymbol{c} \leftarrow \boldsymbol{c} + (R_1 \times \text{DilateRegion}(\boldsymbol{M}_j, n)) \cdot (\boldsymbol{c} == 0)$
    \Else \Comment{Large region detected}
        \State Perform grid-based segmentation: $\boldsymbol{B}_{\text{grid}} \leftarrow \text{SegmentGrid}(\boldsymbol{M}_j, grid\_size)$
        \For{each grid block $B$ in $\boldsymbol{B}_{\text{grid}}$}
            \If{block $B$ contains any positive pixel}
                \State Generate random color $R_2$
                \State Apply color $\boldsymbol{c} \leftarrow \boldsymbol{c} + (R_2 \times B) \cdot (\boldsymbol{c} == 0)$
            \EndIf
        \EndFor
    \EndIf
\EndFor
\State \Return $\boldsymbol{c}$
\end{algorithmic}
\label{alg:01}
\end{algorithm}

\subsection{2D Transformations}
Previous research has shown that increasing the diversity of input images during the generation of adversarial examples enhances both the effectiveness and transferability of the attacks. Techniques such as DIM~\cite{xie2019improving}, TIM~\cite{dong2019evading}, SIM~\cite{lin2019nesterov}, and RSTAM~\cite{liu2022rstam} have introduced methods to manipulate input images, enabling adversarial examples to generalize across different models and tasks.

In this work, we implement the Random Similarity Transformation Strategy from RSTAM~\cite{liu2022rstam,liu2024eap}, which applies random translations, rotations, and scaling to input images. This method ensures that the generated perturbations remain effective against geometric transformations. Additionally, we adopt the Scale-Invariant Strategy from SIM~\cite{lin2019nesterov}, which averages gradients across multiple scaled versions of the image. By integrating these strategies, we improve the robustness and generalization of our adversarial examples, enhancing their transferability across various models and real-world applications.

\subsection{Loss Function}
To guide the adversarial attack, we define the following loss function $ \mathcal{L} $, which aims to reduce the similarity between the adversarial image and the source image while encouraging its similarity to a guidance map:

\begin{equation}
    \mathcal{L} = \frac{<\boldsymbol{y}_a, \boldsymbol{y}_s>}{||\boldsymbol{y}_a||^2 \cdot ||\boldsymbol{y}_s||^2} 
    - \lambda \frac{<\boldsymbol{y}_a, \boldsymbol{y}_g>}{||\boldsymbol{y}_a||^2 \cdot ||\boldsymbol{y}_g||^2}
\end{equation}

where:
\begin{itemize}
    \item $<\cdot,\cdot>$ denotes the inner product of the vectors.
    \item $ \boldsymbol{y}_a = f_{\theta_{I}}\mathcal{F}_{\text{2d}}(\bar{\boldsymbol{x}}) $ denotes the features extracted from the adversarial image $ \bar{\boldsymbol{x}} $ after applying a 2D transformation $ \mathcal{F}_{\text{2d}} $.
    \item $ \boldsymbol{y}_s = f_{\theta_{I}}(\boldsymbol{x}) $ represents the features extracted from the source image $ \boldsymbol{x} $.
    \item $ \boldsymbol{y}_g = f_{\theta_{I}}(\boldsymbol{c})$ corresponds to the features of the guidance map $ \boldsymbol{c}$.
    \item $ f_{\theta_{I}} $ is the image encoder from the SAM used for feature extraction.
    \item $ \lambda $ is a regularization parameter that controls the contribution of the second term. The default value of $ \lambda $ is set to 1. 
\end{itemize}

The loss function $ \mathcal{L} $ drives the adversarial image to move away from the source image while moving closer to the guidance map, allowing for control over the attack's behavior through the regularization parameter $ \lambda $.

The RGA approach enables targeted, region-specific adversarial perturbations that disrupt SAM’s segmentation without requiring external prompts. By focusing on the internal feature structure, RGA achieves a high degree of transferability and robustness against various segmentation models. A detailed algorithmic description of the RGA approach is provided in Algorithm~\ref{alg:RGA}.

\begin{algorithm}[!ht]
\centering
    \caption{Region-Guided Attack (RGA)}

    \begin{algorithmic}[1]
    \Require {Source image $\boldsymbol{x}$.}
    \Require {Segment Anything Model (SAM) $f_{\text{sam}}$; SAM image encoder $f_{\theta_{I}}$.}
    \Require {2D transformation function $\mathcal{F}_{\text{2d}}$; Segmentation and dilation function $\mathcal{F}_{\text{sad}}$.} 
    \Require {Granularity parameter for grid size $\gamma$;
         Number of dilation iterations $n$;
         Perturbation step size $\alpha$;
         Perturbation bound $\epsilon$;
        Decay factor for momentum  $\mu$;
         Number of adversarial iterations $T$.}
     \Require {Loss function $\mathcal{L}$.}
    \Ensure Adversarial perturbation: $\delta$.
    
    \State Initialize the gradient: $\boldsymbol{g}_0 \leftarrow 0$
    
    \State Initialize the adversarial image with uniform noise:
        $\bar{\boldsymbol{x}} \leftarrow \boldsymbol{x} + \mathcal{U}(-\epsilon, \epsilon)$

    \State Query SAM with the source image: $\mathcal{M} \leftarrow f_{\text{sam}}(\boldsymbol{x})$
    
    \State Generate the guidance map using the SAD strategy: $\boldsymbol{c} \leftarrow \mathcal{F}_{\text{sad}}(\mathcal{M}; \gamma, n)$
    
    \For{$t = 0$ to $T-1$} 
        \State Compute the gradient of the loss function:
            \begin{equation*}
                \boldsymbol{g}_{t}^{'} \leftarrow \nabla_{\bar{\boldsymbol{x}}_{t}} \mathcal{L} \big( f_{\theta_{I}}(\boldsymbol{x}), f_{\theta_{I}} \mathcal{F}_{\text{2d}}(\bar{\boldsymbol{x}}_{t}), f_{\theta_{I}}(\boldsymbol{c}) \big)
            \end{equation*}
        \State Update the gradient using momentum:
        \begin{equation*}
            \boldsymbol{g}_{t+1} \leftarrow \mu \cdot \boldsymbol{g}_{t} + \boldsymbol{g}_{t}^{'}
        \end{equation*}
        \State Update the adversarial image using the sign of the gradient:
        \begin{equation*}
            \bar{\boldsymbol{x}}_{t+1} \leftarrow \mathop{\mathrm{Clip}}_{[\boldsymbol{x}-\epsilon, \boldsymbol{x}+\epsilon]}
            \big( \bar{\boldsymbol{x}}_t + \alpha \cdot \mathrm{sign}(\boldsymbol{g}_{t+1}) \big)
        \end{equation*}
    \EndFor
    \State Compute the adversarial perturbation: $\delta \leftarrow \bar{\boldsymbol{x}}_{T}-\boldsymbol{x}$
    \State \Return $\delta$
    \end{algorithmic}
\label{alg:RGA}
\end{algorithm}

\section{Experiments}\label{sec5}
\subsection{Experiment Settings}
 \textbf{Datasets:} For evaluation, we selected the SAM-1B dataset~\cite{kirillov2023segment}, which comprises a diverse and extensive collection of images that closely resemble real-world scenarios, including various environments, object types, and lighting conditions. For our study, we focused on a subset of the first 1,000 images, labeled from sa\_1.jpg to sa\_1000.jpg. This carefully chosen subset includes a total of 98,875 masks, representing a substantial quantity that enhances the robustness and statistical significance of our findings.

 \textbf{Compared Baselines:}
To assess the effectiveness of our proposed method, we conducted comparisons with several established adversarial attack techniques on segmentation models. The baselines included:
\begin{enumerate}
\item Attack-SAM-K (ASK)~\cite{zhang2023ASK}: Utilizes a large number of prompts to globally alter SAM’s feature responses, challenging its segmentation robustness.

\item Transferable Adversarial Perturbations (TAP)~\cite{zhou2018TAP}: Focuses on generating adversarial perturbations with high transferability across different models, exposing general model vulnerabilities.

\item Intermediate-Level Perturbation Decay (ILPD)~\cite{li2023ILPD}: Extends TAP by increasing perturbation magnitude at intermediate feature levels, further testing model resilience.

\item Activation Attack (AA)~\cite{Inkawhich2019AA}: Aligns adversarial features with target features, offering precise control over model output alterations.

\item Prompt-Agnostic Target Attack (PATA)~\cite{zheng2023PATA}: Enhances adversarial feature dominance without reliance on specific prompts, making it adaptable to varied input conditions.

\item PATA++~\cite{zheng2023PATA}: An improved version of PATA that dynamically adjusts competition images, optimizing adversarial effects through iterative updates.

\item Unsegment Anything by Simulating Deformation (UAD)~\cite{lu2024UAD}: Alters image structure to disrupt segmentation, leveraging deformation to enhance adversarial robustness and transferability.
\end{enumerate}

 \textbf{Evaluation Metrics:} To assess the performance of our adversarial attack method, we employed the following evaluation metrics:

Mean Intersection over Union (mIoU): The mIoU measures the average overlap between the predicted segmentation and the ground truth masks. It is calculated by taking the mean of the Intersection over Union (IoU) across all masks. The mIoU is defined as:
    \begin{equation}
        \text{mIoU} = \frac{1}{N} \sum_{m=1}^N \frac{|P_m \cap G_m|}{|P_m \cup G_m|}
    \end{equation}
where: $ N $ is the total number of masks, $ P_m $ is the predicted segmentation for mask $ m $, and $ G_m $ is the corresponding ground truth segmentation for mask $ m $.
    
Attack Success Rate at IoU $ \leq 50\% $ (ASR@50): ASR@50 measures the proportion of SAM-generated masks, across all adversarial examples, where the IoU with the ground truth mask is less than or equal to 50\%. This metric evaluates the effectiveness of the attack by indicating how frequently it significantly reduces segmentation accuracy. The formula for ASR@50 is:
    \begin{equation}
        \text{ASR@50} = \frac{1}{N} \sum_{m=1}^N \mathbb{I} (\text{IoU}_m \leq 0.50)
    \end{equation}
    where $ N $ is the total number of masks across all adversarial examples,
 $ \text{IoU}_m $ is the IoU for mask $ m $, and $ \mathbb{I} $ is the indicator function, equal to 1 if $ \text{IoU}_m \leq 0.50 $, and 0 otherwise.

Attack Success Rate at IoU $ \leq 10\% $ (ASR@10): Similar to ASR@50, ASR@10 measures the proportion of masks for which the IoU with the ground truth mask is less than or equal to 10\%, across all adversarial examples. This metric captures cases where the attack is highly effective, causing significant degradation in segmentation accuracy. The formula for ASR@10 is:
    \begin{equation}
        \text{ASR@10} = \frac{1}{N} \sum_{m=1}^N \mathbb{I} (\text{IoU}_m \leq 0.10)
    \end{equation}
where the terms $ N $, $ \text{IoU}_m $, and $ \mathbb{I} $ are as defined above, with the threshold adjusted to $ 10\% $.

 \textbf{Implementation Details:} In our experiments, we set the following default parameters. The granularity parameter for grid size, $\gamma$, is set to 0.1 to ensure adequate detail in the segmentation process, and the number of dilation iterations, $n$, is set to 100 to expand regions effectively. For all adversarial attacks, we set the perturbation bound, $\epsilon$, to 8/255 and the perturbation step size, $\alpha$, to 2/255, with the number of adversarial iterations, $T$, set to 40 to balance effectiveness and computational efficiency. All experiments are conducted on four NVIDIA GeForce GTX 3090 GPUs, and the models are implemented using the PyTorch~\cite{NEURIPS2019_9015} framework.

\begin{table*}[!ht]
\centering
\caption{Quantitative comparison of different adversarial attack methods on various segmentation models. The metrics include mean Intersection over Union (mIoU), Attack Success Rate at thresholds 50\% (ASR@50), and 10\% (ASR@10). Lower mIoU and higher ASR values indicate better attack performance. The best performances in each block are shown in \textbf{bold}.}
\label{tab:01}
\resizebox{\textwidth}{!}{%
\begin{tabular}{@{}c|c|c|c|c|c|c|c|c|c|c|c|c@{}}
\toprule
\multirow{2}{*}{Method} & \multicolumn{3}{c|}{SAM-B(white-box)} & \multicolumn{3}{c|}{SAM-L}        & \multicolumn{3}{c|}{SAM-H}          & \multicolumn{3}{c}{FastSAM~\cite{zhao2023fast}}       \\ \cmidrule(l){2-13} 
                        & mIoU↓           & ASR@50↑  & ASR@10↑  & mIoU↓         & ASR@50↑ & ASR@10↑ & mIoU↓           & ASR@50↑ & ASR@10↑ & mIoU↓         & ASR@50↑ & ASR@10↑ \\ \midrule
ASK~\cite{zhang2023ASK}  &  $68.07\pm28.65$ & $24.10$& $6.87$ & $77.14\pm25.05$ & $14.46$ & $4.13$ & $78.71\pm24.02$ & $12.93$ & $3.51$ & $38.13\pm40.66$ & $59.90$ & $48.43$\\
TAP~\cite{zhou2018TAP}  &  $63.49\pm32.58$ & $29.69$ & $13.12$ & $75.12\pm27.83$ & $16.96$ & $6.68$ & $77.36\pm26.21$ & $14.55$ & $5.31$ & $37.67\pm40.89$ & $60.53$ & $49.45$\\
ILPD~\cite{li2023ILPD}  & $63.21\pm32.54$ & $30.15$ & $13.09$ & $75.17\pm27.75$ & $16.82$ & $6.59$ & $77.52\pm26.02$ & $14.37$ & $5.18$ & $37.84\pm40.84$ & $60.30$ & $49.02$\\
AA~\cite{Inkawhich2019AA}  & $61.06\pm32.33$ & $32.48$ & $13.11$ & $70.70\pm29.74$ & $21.49$ & $8.67$ & $72.87\pm28.61$ & $19.13$ & $7.39$ & $32.64\pm39.58$ & $65.86$ & $55.10$\\
PATA~\cite{zheng2023PATA}  & $61.36\pm32.31$ & $32.23$ & $13.04$ & $70.81\pm29.73$ & $21.27$ & $8.62$ & $73.07\pm28.49$ & $18.66$ & $7.30$ & $32.74\pm39.56$ & $65.66$ & $54.97$\\
PATA++~\cite{zheng2023PATA}  & $61.54\pm32.22$ & $32.00$ & $12.94$ & $71.02\pm29.55$ & $21.16$ & $8.38$ & $73.16\pm28.44$ & $18.84$ & $7.22$ & $32.85\pm39.60$ & $65.69$ & $54.65$\\
UAD~\cite{lu2024UAD}     & $51.53\pm34.00$ & $43.89$ & $20.79$ & $66.07\pm32.04$ & $26.44$ & $12.27$ & $68.96\pm30.87$ & $23.42$ & $10.23$ & $28.83\pm38.36$ & $69.95$ & $59.63$\\ 
\midrule
RGA(Ours)  & $\mathbf{26.64\pm31.95}$  & $\mathbf{72.08}$ & $\mathbf{55.22}$  & $\mathbf{28.63\pm32.59}$  & $\mathbf{69.93}$ & $\mathbf{52.09}$   &  $\mathbf{30.98\pm33.55}$  & $\mathbf{67.38}$ & $\mathbf{48.82}$ & $\mathbf{5.32\pm14.98}$ & $\mathbf{96.91}$ & $\mathbf{89.56}$   \\
\bottomrule
\end{tabular}%
}
\end{table*}

\subsection{Quantitative Evaluation}
We evaluate the performance of our proposed RGA method against several state-of-the-art adversarial attack techniques, including ASK~\cite{zhang2023ASK}, TAP~\cite{zhou2018TAP}, ILPD~\cite{li2023ILPD}, AA~\cite{Inkawhich2019AA}, PATA~\cite{zheng2023PATA}, PATA++\cite{zheng2023PATA}, and UAD\cite{lu2024UAD}. The evaluation metrics used are mean Intersection over Union (mIoU) (lower is better), Attack Success Rate at thresholds 50\% (ASR@50), and 10\% (ASR@10) (higher is better). The experiments are conducted on four models: SAM-B (white-box), SAM-L, SAM-H, and FastSAM~\cite{zhao2023fast}. The results are summarized in Table~\ref{tab:01}.

As shown in Table~\ref{tab:01}, our RGA method significantly outperforms existing adversarial attack methods across all models and evaluation metrics. Specifically, on the SAM-B (white-box) model, RGA achieves a mIoU of $26.64\pm31.95$, which is substantially lower than the closest competitor, UAD, with a mIoU of $51.53\pm34.00$. This indicates that RGA is more effective in degrading the segmentation quality of the model. Additionally, RGA attains the highest ASR@50 and ASR@10 rates of $72.08\%$ and $55.22\%$, respectively, demonstrating superior attack success.

Similar trends are observed on the SAM-L and SAM-H models, where RGA consistently achieves lower mIoU values and higher ASR rates compared to other methods. Notably, on the FastSAM model, RGA achieves an exceptionally low mIoU of $5.32\pm14.98$ and ASR@50 and ASR@10 rates of $96.91\%$ and $89.56\%$, respectively. This highlights the effectiveness of RGA in both white-box and black-box settings.

The superior performance of RGA can be attributed to its ability to generate regionally optimized adversarial perturbations that effectively disrupt the segmentation process of the models. These results confirm the efficacy of our approach in compromising the robustness of segmentation models to adversarial attacks.

\begin{figure}[!ht]
    \centering
    \includegraphics[width=1\linewidth]{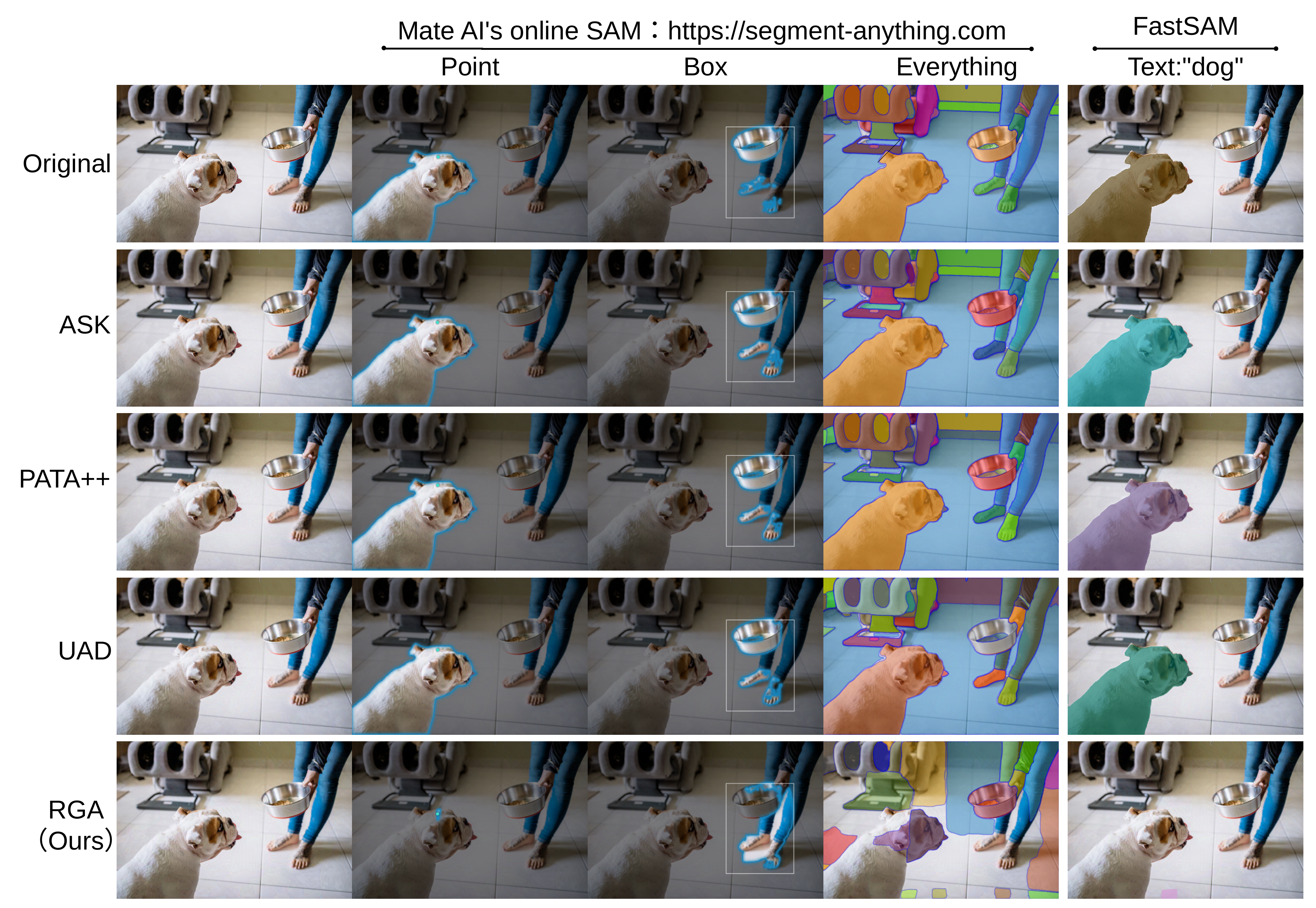}
    \caption{Qualitative comparison of segmentation results under different adversarial attacks on Meta AI’s online SAM and FastSAM models in a black-box setting. The comparison includes various prompts, with Point, Box, and Everything on Meta AI's online SAM, and Text (``dog'') on FastSAM. The perturbation bound $\epsilon$ is reduced to $4/255$. While most other attack methods are largely ineffective, our method continues to successfully disrupt segmentation. }
    \label{fig:qual}
\end{figure}
\begin{figure}[!ht]
    \centering
    \includegraphics[width=1\linewidth]{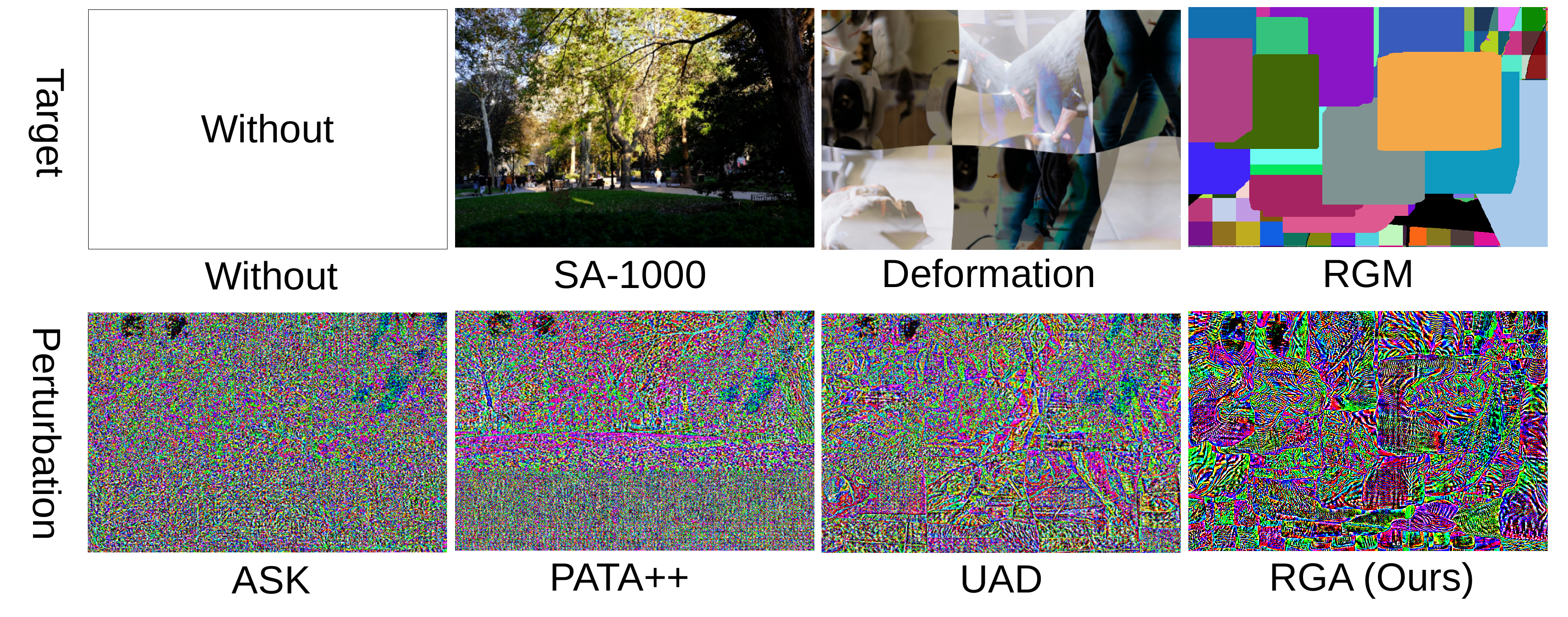}
    \caption{Visualizations of generated adversarial perturbations from different methods, corresponding to different targets used during the attack on segmentation models.}
    \label{fig:perturbation}
\end{figure}

\subsection{Qualitative Evaluation}

We conducted qualitative evaluations to assess the effectiveness of the proposed RGA in a black-box setting, specifically targeting Meta AI's online SAM model and FastSAM. In our experiments, we validated the attack on Meta AI's online SAM model using various prompt types: Point, Box, and Everything. For FastSAM, a Text prompt set to ``dog'' guided the segmentation process.

In previous experiments, we used a default perturbation bound of $\epsilon = 8/255$. In this qualitative evaluation, however, for all attack methods, the perturbation bound $\epsilon$ is reduced to $4/255$ to ensure a more challenging and subtle attack scenario, allowing us to better compare the effectiveness of each method under constrained conditions.

Figure~\ref{fig:qual} presents visual results comparing the segmentation masks generated under normal conditions and under adversarial attacks by RGA, ASK, PATA++, and UAD. The comparison illustrates how RGA effectively disrupts segmentation performance across all prompt types on Meta AI's online SAM, producing segmentation errors by fragmenting large regions and expanding small regions. This validates RGA’s capability to induce substantial segmentation errors while maintaining visually subtle perturbations to the input image.

In comparison, FastSAM’s segmentation result using the Text prompt “dog” demonstrates the versatility and transferability of RGA in handling different prompt-based black-box models. Our results indicate that RGA consistently outperforms existing adversarial attacks in reducing segmentation quality across both Meta AI’s online SAM and FastSAM models.

Figure~\ref{fig:perturbation} further illustrates the effectiveness of RGA by visualizing the adversarial perturbations generated according to the RGM. The RGM acts as the target for RGA, guiding the perturbations to specifically manipulate segmented regions identified during the attack. By comparing the perturbations applied by RGA against those from ASK, PATA++, and UAD, we can observe distinct patterns that reveal how each method influences segmentation outputs. RGA’s use of the RGM allows for strategic alterations—fragmenting large segments and expanding small ones—thereby reinforcing its ability to induce significant segmentation errors. This targeted approach, depicted in Figure~\ref{fig:perturbation}, underscores RGA’s robustness in maintaining subtle visual integrity while effectively disrupting segmentation performance across various conditions. Together, these figures highlight RGA's capability to perform sophisticated adversarial attacks tailored to the structural characteristics of segmentation models.

\begin{table*}[!ht]
\centering
\caption{Ablation study results for the RGA on SAM-B (white-box), SAM-L, and SAM-H models. Each row shows the effect of enabling different components of the RGA framework, including Region-Guided Map (RGM), Momentum Iteration (MI), Random Similarity Transformation (RST), and Scale-Invariance (SI). The best performances in each block are shown in \textbf{bold}.}
\label{tab:ablation}
\resizebox{\textwidth}{!}{%
\begin{tabular}{cccc|ccc|ccc|ccc}
\toprule
\multirow{2}{*}{RGM} & \multirow{2}{*}{MI} & \multirow{2}{*}{RST} & \multirow{2}{*}{SI} & \multicolumn{3}{c|}{SAM-B(white-box)} & \multicolumn{3}{c|}{SAM-L} & \multicolumn{3}{c}{SAM-H} \\
\cmidrule(lr){5-7} \cmidrule(lr){8-10} \cmidrule(lr){11-13}
 &  &  &  & mIoU↓ & ASR@50↑ & ASR@10↑ & mIoU↓ & ASR@50↑ & ASR@10↑ & mIoU↓ & ASR@50↑ & ASR@10↑ \\
\midrule
\xmark & \xmark & \xmark & \xmark & 49.98 & 46.82 & 23.66 & 71.49 & 21.15 & 7.92 & 75.24 & 16.90 & 5.41 \\
\cmark & \xmark & \xmark & \xmark & 45.71 & 50.67 & 32.83 & 70.66 & 22.14 & 9.30 & 74.54 & 17.42 & 6.85 \\
\cmark & \cmark & \xmark & \xmark & 46.63 & 49.77 & 31.67 & 69.98 & 22.57 & 10.19 & 73.89 & 17.76 & 7.42 \\
\cmark & \cmark & \cmark & \xmark & 29.71 & 68.84 & 52.31 & 33.38 & 64.75 & 47.53 & 38.11 & 59.57 & 41.57 \\
\xmark & \cmark & \cmark & \cmark & 30.27 & 71.59 & 40.39 & 32.02 & 69.40 & 38.14 & 33.22 & 68.04 & 36.76 \\
\cmark & \cmark & \cmark & \cmark & \textbf{26.87} & \textbf{72.99} & \textbf{55.60} & \textbf{28.27} & \textbf{69.64} & \textbf{52.66} & \textbf{31.15} & \textbf{67.41} & \textbf{48.81} \\
\bottomrule
\end{tabular}
}
\end{table*}
\subsection{Ablation Studies}
To comprehensively evaluate the contributions of various components in the RGA framework, we conduct a series of ablation studies focusing on three main aspects: the individual impact of the proposed components, the integration of RGA with traditional adversarial attack methods, and the evaluation of different adversarial target types. The ablation experiments are conducted on SAM-B (white-box), SAM-L, and SAM-H models, and the results are presented in three tables.

\subsubsection{Component Analysis}
To better understand the impact of different components in our RGA framework, we evaluate the following: Region-Guided Map (RGM), Momentum Iteration (MI), Random Similarity Transformation (RST), and Scale-Invariance (SI). Table~\ref{tab:ablation} provides a summary of the results, highlighting the effectiveness of each component when applied individually or in combination.

As observed in Table~\ref{tab:ablation}, the baseline version of RGA, where none of the components (RGM, MI, RST, SI) are applied, demonstrates relatively low Attack Success Rates (ASR) and high mIoU values, indicating limited effectiveness in degrading the segmentation performance of the SAM models. Specifically, the baseline achieves a mIoU of $49.98$ for SAM-B, $71.49$ for SAM-L, and $75.24$ for SAM-H, with low ASR@10 values.

Adding the RGM significantly improves the performance, resulting in decreased mIoU and increased ASR@50 and ASR@10 across all models. Incorporating Momentum Iteration (MI) and Random Similarity Transformation (RST) further enhances the attack success, achieving a reduction in mIoU to $29.71$ for SAM-B and $33.38$ for SAM-L, accompanied by notable improvements in ASR values. The final configuration, where all components (RGM, MI, RST, SI) are included, achieves the best performance, with SAM-B's mIoU dropping to $26.87$, and ASR@50 and ASR@10 reaching $72.99\%$ and $55.60\%$, respectively. This demonstrates the synergy between the components and their collective contribution to the overall performance of RGA.

\begin{table*}[!ht]
\centering
\caption{Impact of integrating the Region-Guided Map (RGM) with traditional adversarial attack methods. The best performances in each block are shown in \textbf{bold}.}
\label{tab:imp}
\resizebox{\textwidth}{!}{%
\begin{tabular}{c|ccc|ccc|ccc}
\toprule
\multirow{2}{*}{Method} & \multicolumn{3}{c|}{SAM-B(white-box)} & \multicolumn{3}{c|}{SAM-L} & \multicolumn{3}{c}{SAM-H} \\
\cmidrule(lr){2-4} \cmidrule(lr){5-7} \cmidrule(lr){8-10}
 & mIoU↓ & ASR@50↑ & ASR@10↑ & mIoU↓ & ASR@50↑ & ASR@10↑ & mIoU↓ & ASR@50↑ & ASR@10↑ \\
\midrule
MIM~\cite{dong2018boosting} & $54.64$ & 40.92 & 19.01 & $72.27$ & 20.09 & 7.45 & $75.44$ & 16.69 & 5.52 \\
MIM+RGM & \textbf{46.59} & \textbf{49.70} & \textbf{32.04} & \textbf{70.09} & \textbf{22.20} & \textbf{10.05} & \textbf{74.13} & \textbf{17.36} & \textbf{7.05} \\ \midrule
DIM~\cite{xie2019improving} & 33.46 & 67.95 & 36.16 & 44.23 & 53.85 & 26.45 & 47.79 & 50.13 & 24.04 \\
DIM+RGM & \textbf{28.32} & \textbf{70.04} & \textbf{54.14} & \textbf{40.51} & \textbf{56.50} & \textbf{39.93} &\textbf{ 46.34 }& \textbf{49.89} & \textbf{32.96} \\
\bottomrule
\end{tabular}
}
\end{table*}

\subsubsection{Integrating RGM with Traditional Methods}
We further evaluate the impact of integrating the RGM component with traditional adversarial attack methods, such as MIM~\cite{dong2018boosting} and DIM~\cite{xie2019improving}. Table~\ref{tab:imp} presents the results, demonstrating the effectiveness of RGM in enhancing the transferability and robustness of these traditional methods.

From Table~\ref{tab:imp}, it is evident that adding RGM to both MIM and DIM results in substantial performance gains across all metrics on SAM-B (white-box), SAM-L, and SAM-H models. For example, the mIoU for MIM on SAM-B decreases from $54.64$ to $46.59$ when integrated with RGM, while ASR@50 and ASR@10 improve from $40.92\%$ to $49.70\%$ and $19.01\%$ to $32.04\%$, respectively. Similarly, DIM+RGM achieves a reduction in mIoU to $28.32$ and improvements in ASR metrics, underscoring the effectiveness of incorporating RGM to enhance traditional attack approaches.

\begin{table*}[!ht]
\centering
\caption{Comparison of different target types for adversarial perturbations, including black, white, random noise, and samples from the SA-1000 dataset. The best performances in each block are shown in \textbf{bold}.}
\label{tab:target}
\resizebox{\textwidth}{!}{%
\begin{tabular}{c|ccc|ccc|ccc}
\toprule
\multirow{2}{*}{Target} & \multicolumn{3}{c|}{SAM-B(white-box)} & \multicolumn{3}{c|}{SAM-L} & \multicolumn{3}{c}{SAM-H} \\
\cmidrule(lr){2-4} \cmidrule(lr){5-7} \cmidrule(lr){8-10}
 & mIoU↓ & ASR@50↑ & ASR@10↑ & mIoU↓ & ASR@50↑ & ASR@10↑ & mIoU↓ & ASR@50↑ & ASR@10↑ \\
\midrule
black & 34.17 & 59.90 & 51.44 & 40.18 & 54.79 & 42.24 & 44.60 & 50.18 & 36.34 \\
white & 33.10 & 61.00 & 52.90 & 39.16 & 55.89 & 43.78 & 44.17 & 50.95 & 36.60 \\
random noise & 32.15 & 65.65 & 46.73 & 37.22 & 58.75 & 39.96 & 42.28 & 54.91 & 31.36 \\
SA-1000 & 30.05 & 68.71 & 48.64 & 34.27 & 63.62 & 43.59 & 36.89 & 60.85 & 40.79 \\
RGM & \textbf{26.87} & \textbf{72.99} & \textbf{55.60} & \textbf{28.27} & \textbf{69.64} & \textbf{52.66} & \textbf{31.15} & \textbf{67.41} & \textbf{48.81} \\
\bottomrule
\end{tabular}
}
\end{table*}

\subsubsection{Evaluating Different Target Types}
To further evaluate the versatility of our RGA framework, we conduct experiments using different target types for adversarial perturbations, including black, white, random noise, and randomly selected samples from the SA-1000 dataset. The results are summarized in Table~\ref{tab:target}, which shows that our RGM approach consistently outperforms other target types in terms of mIoU and ASR metrics across all models.

As shown in Table~\ref{tab:target}, RGM achieves the lowest mIoU and highest ASR values across SAM-B, SAM-L, and SAM-H models. For example, on the SAM-B (white-box) model, RGM achieves a mIoU of $26.87$, which is significantly better than the other target types, such as SA-1000, which yields a mIoU of $30.05$. The ASR@50 and ASR@10 values for RGM are also notably higher, demonstrating the superior ability of RGM to degrade segmentation quality effectively.

In summary, the ablation studies conducted across the three experiments highlight the importance of each component in the RGA framework, the effectiveness of integrating RGM with traditional adversarial methods, and the superior performance of RGM over other target types. The results validate the strength of our approach in improving the transferability and effectiveness of adversarial attacks against segmentation models.




\subsection{Sensitivity Analysis of Hyper-parameters}
In this section, we conduct a sensitivity analysis to evaluate the impact of key hyper-parameters on the performance of our RGA framework. The hyper-parameters examined include the perturbation bound $ \epsilon $, the number of adversarial iterations $ T $, and the parameters associated with the Segmentation and Dilation Strategy, specifically the granularity parameter for grid size $ \gamma $ and the number of dilation iterations $ n $.

\begin{figure*}[!ht]
    \centering
    \subfigure[mIoU of the target model SAM-B]{
		\includegraphics[width=0.45\linewidth]{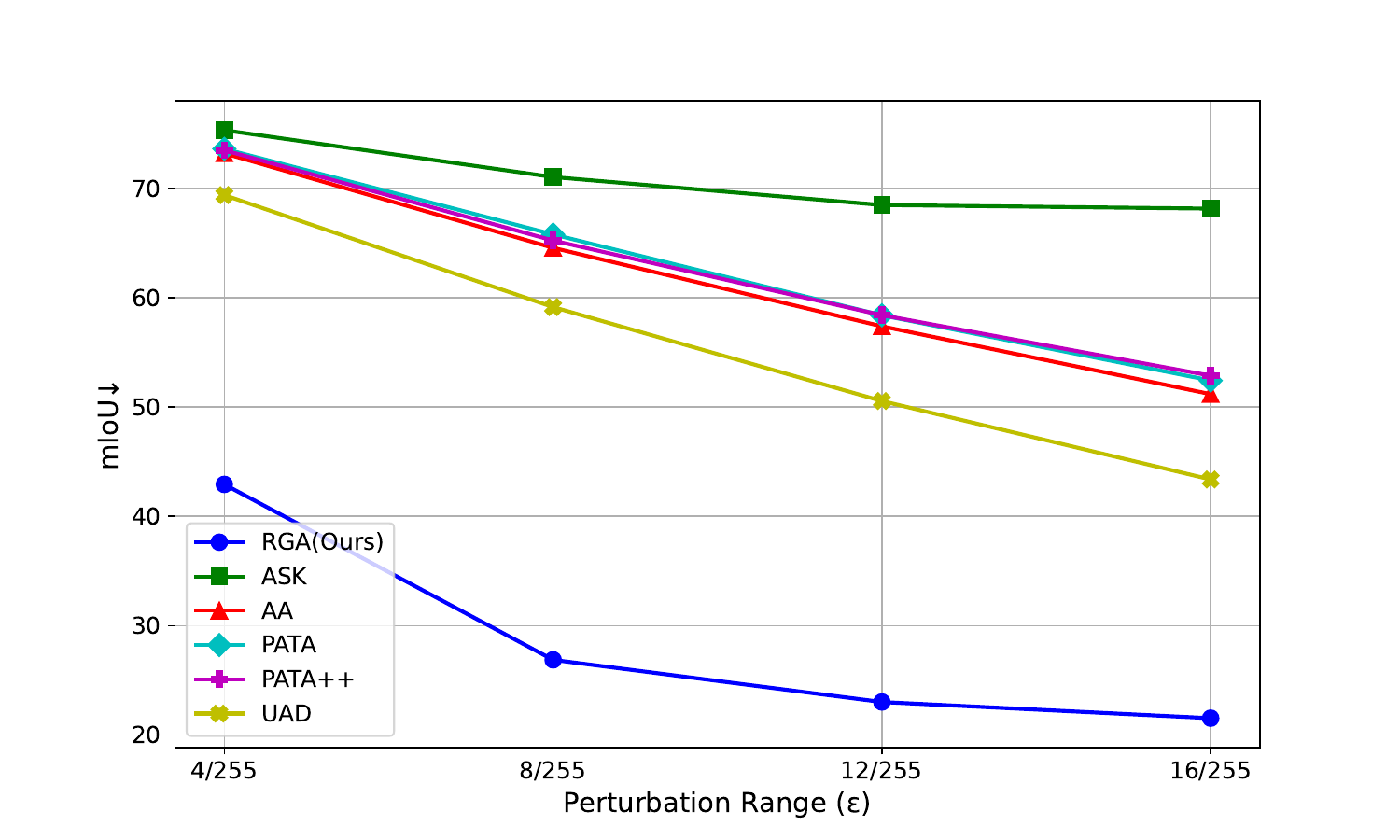}}
    \subfigure[ASR of the target model SAM-B]{
		\includegraphics[width=0.45\linewidth]{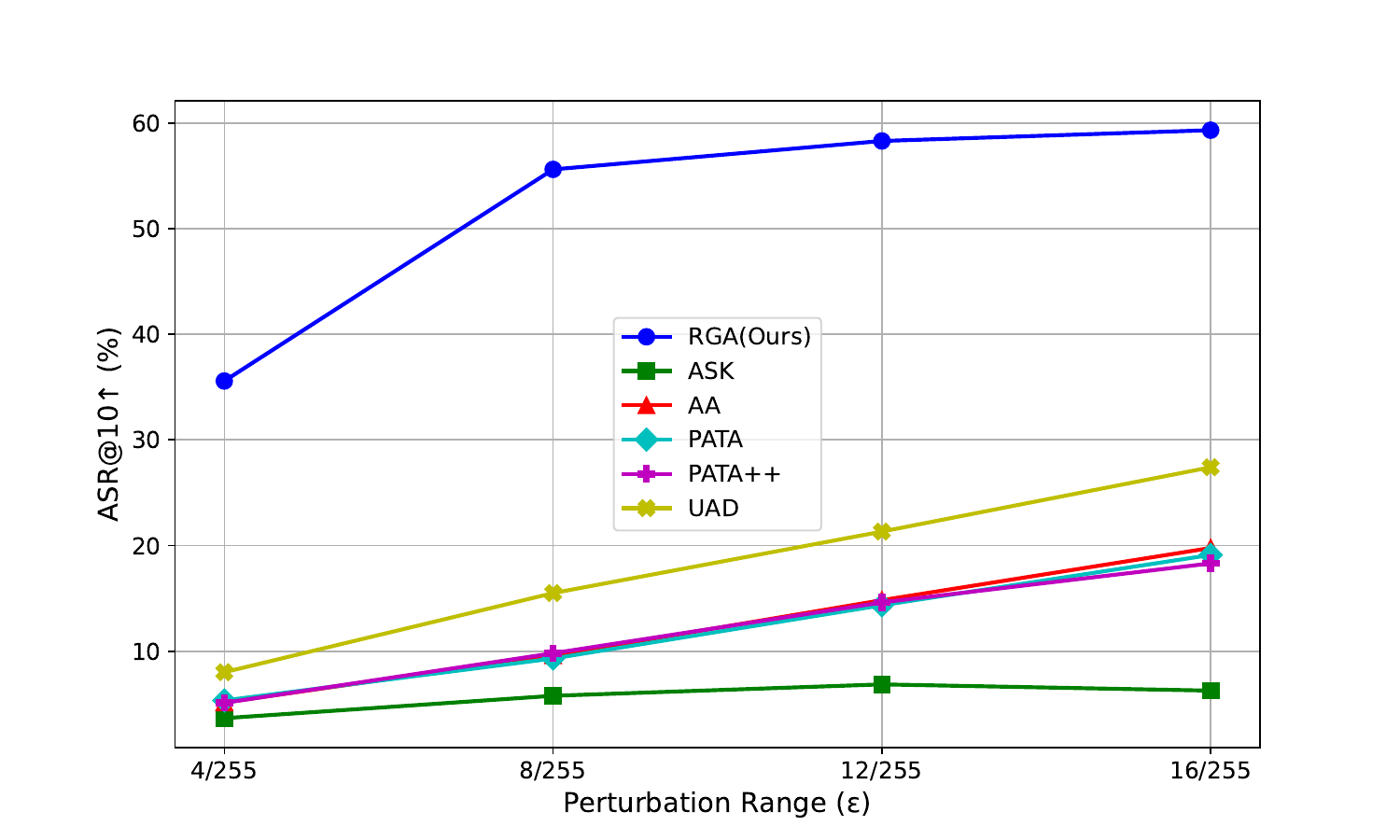}}
    \subfigure[mIoU of the target model SAM-H]{
		\includegraphics[width=0.45\linewidth]{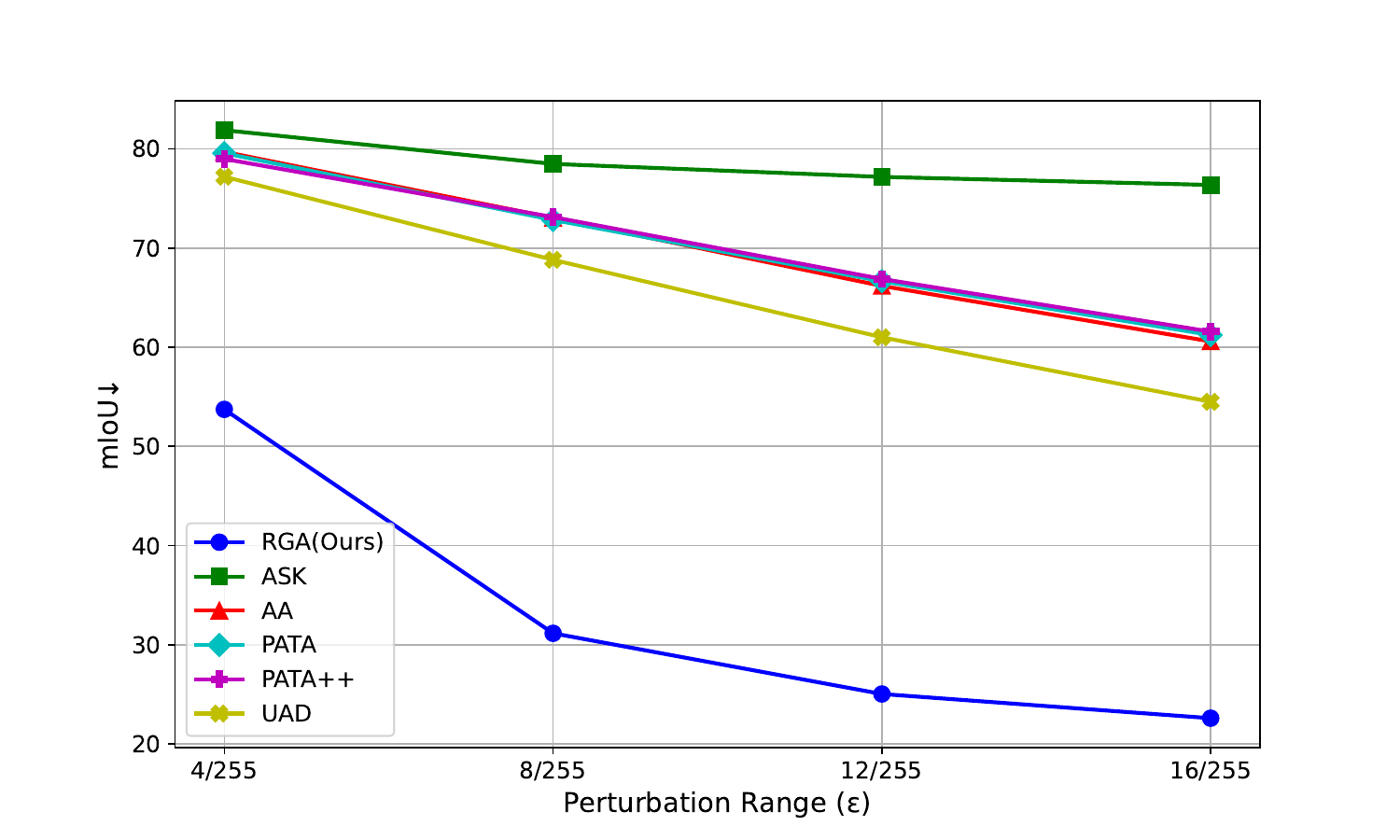}}
  \subfigure[ASR of the target model SAM-H]{
		\includegraphics[width=0.45\linewidth]{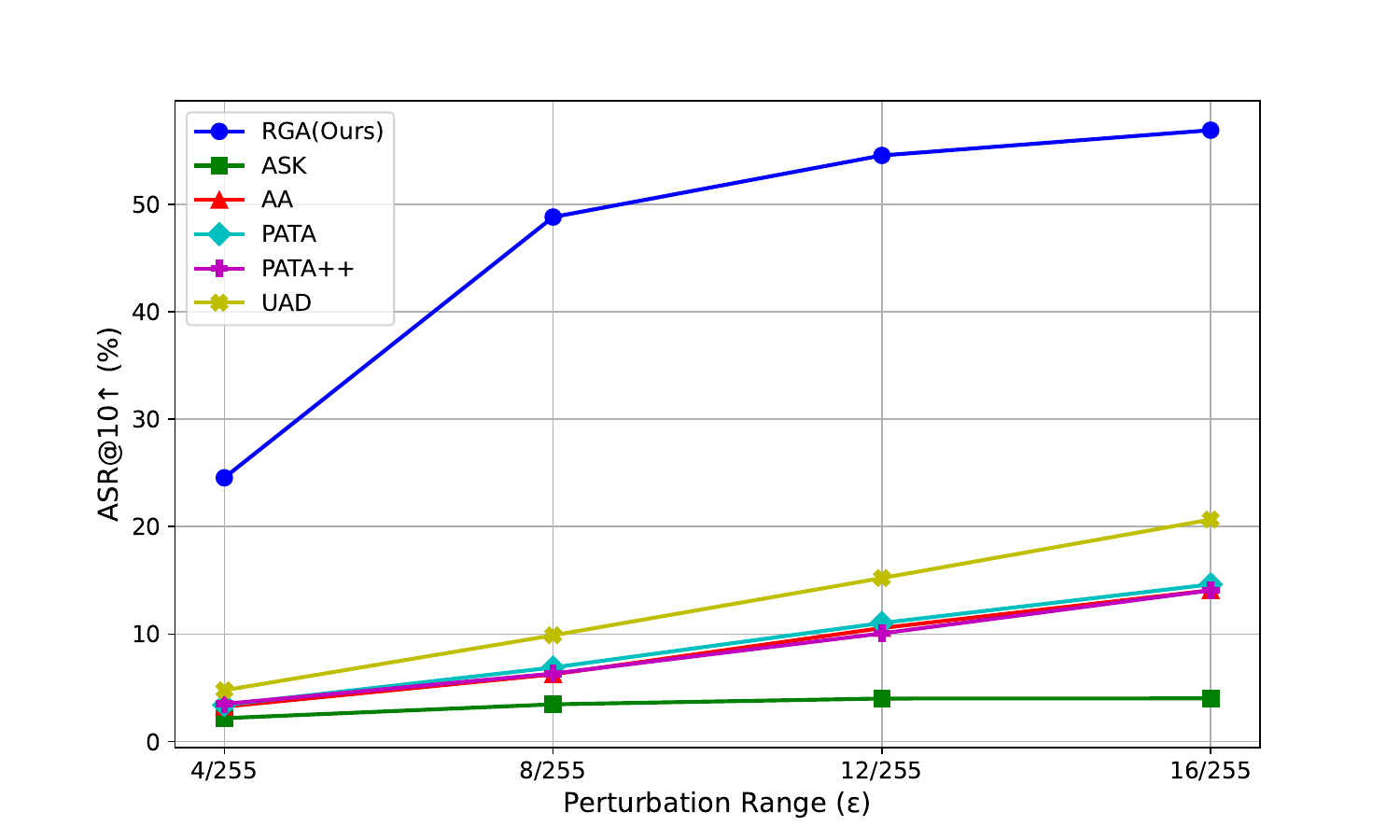}}
    \caption{Sensitivity analysis results of the perturbation bound, $\epsilon$.}
    \label{fig:epsilon}
\end{figure*}
\subsubsection{Perturbation Bound $ \epsilon $}
The perturbation bound $ \epsilon $ plays a critical role in defining the maximum allowable distortion applied to the input images. As shown in Figure~\ref{fig:epsilon}, we evaluate various values of $ \epsilon $ to determine how the extent of perturbation affects the attack's effectiveness. Lower values of $ \epsilon $ generally result in more subtle perturbations, which can help maintain the perceptual quality of the input while still achieving a significant degradation in segmentation performance. Conversely, higher values of $ \epsilon $ can lead to more aggressive attacks, potentially compromising the model's integrity more effectively but also risking the visibility of the perturbations.

\begin{figure*}[!ht]
    \centering
    \subfigure[mIoU of the target model SAM-B]{
		\includegraphics[width=0.45\linewidth]{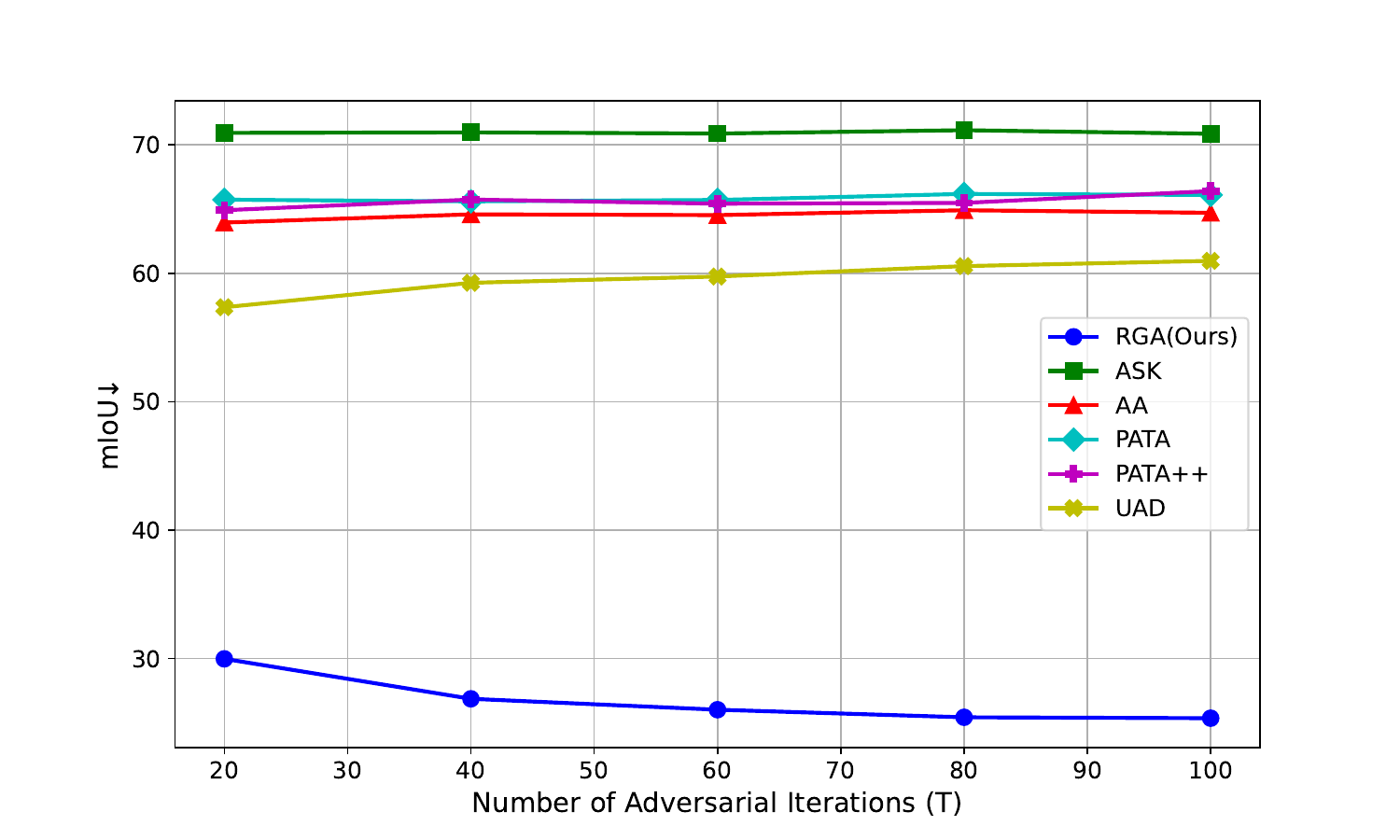}}
    \subfigure[ASR of the target model SAM-B]{
		\includegraphics[width=0.45\linewidth]{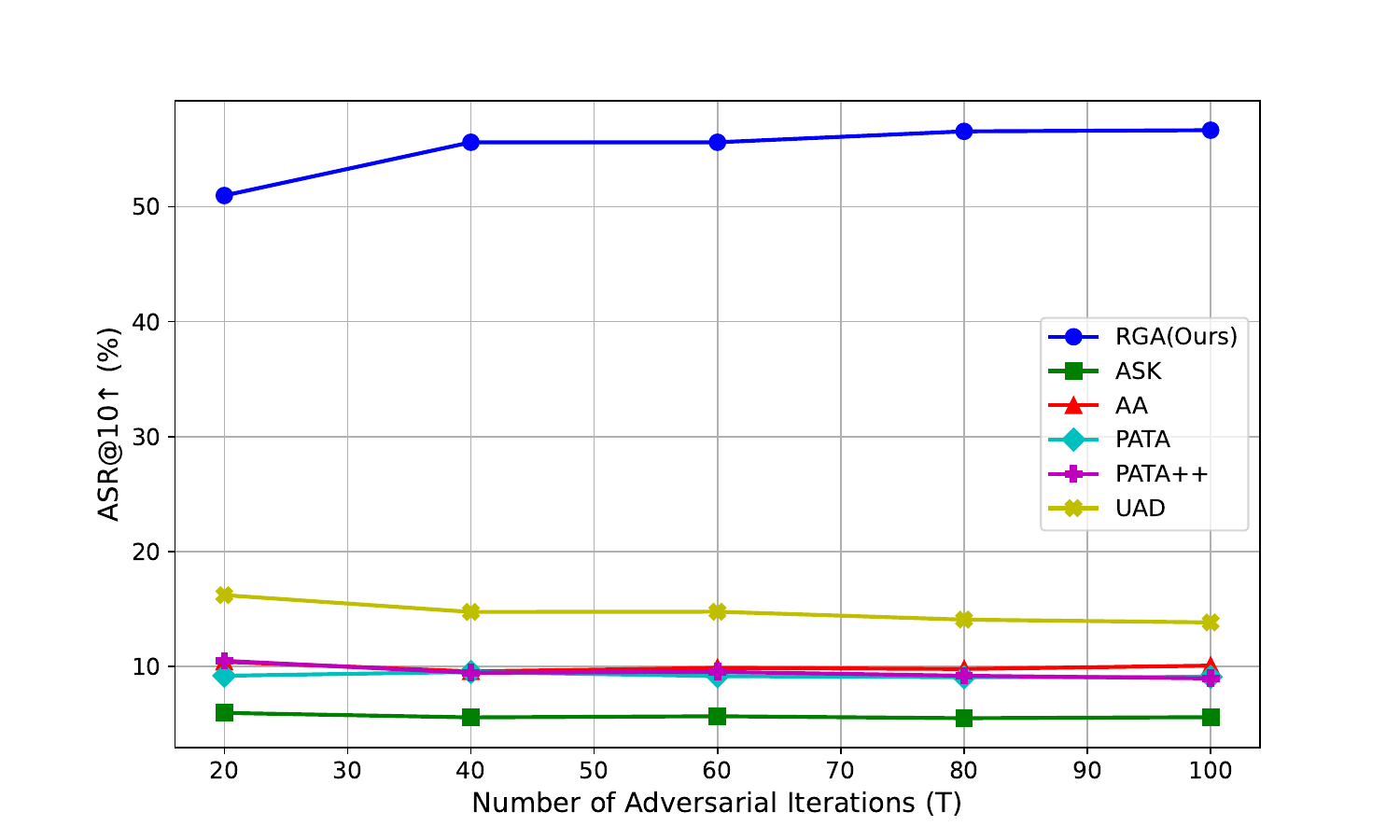}}
    \subfigure[mIoU of the target model SAM-H]{
		\includegraphics[width=0.45\linewidth]{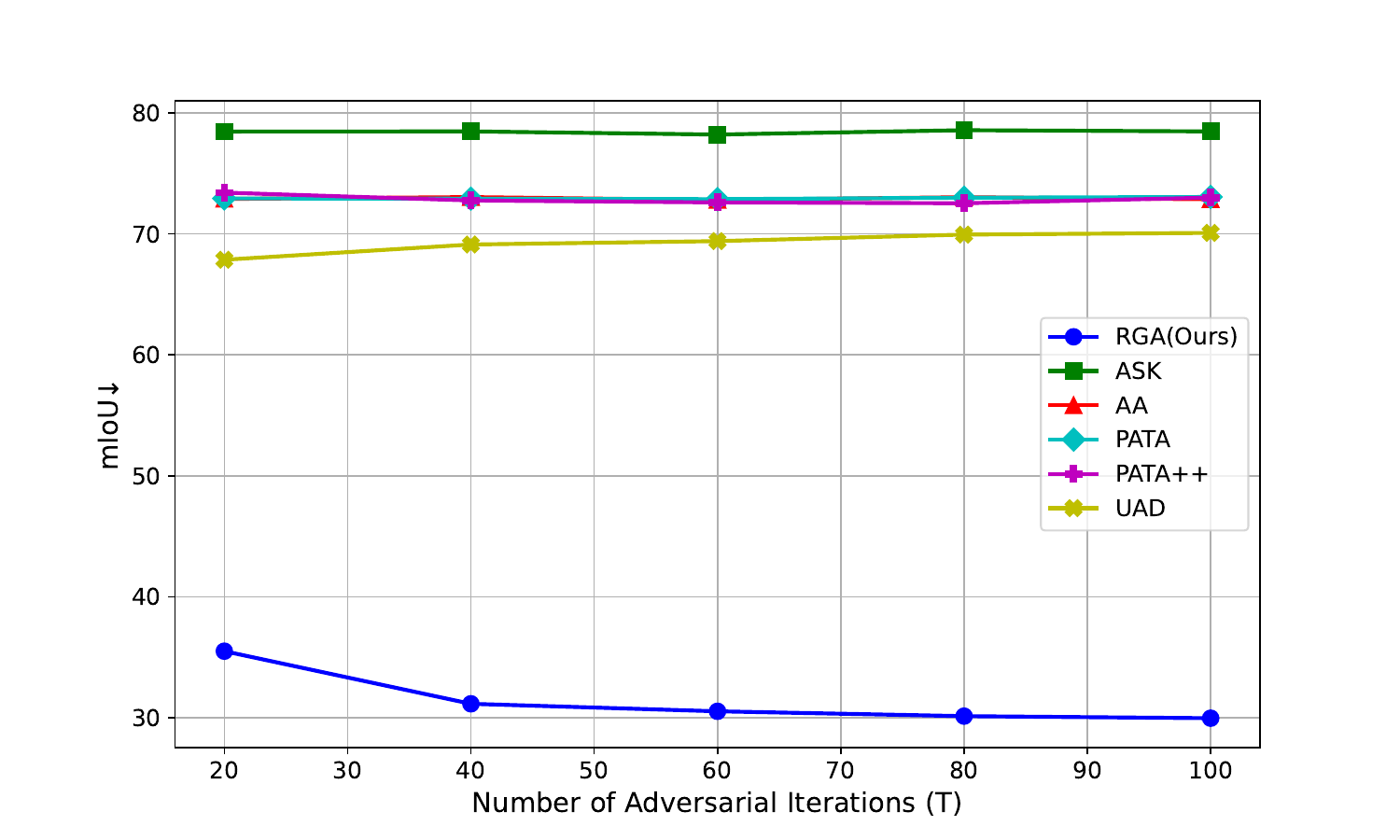}}
  \subfigure[ASR of the target model SAM-H]{
		\includegraphics[width=0.45\linewidth]{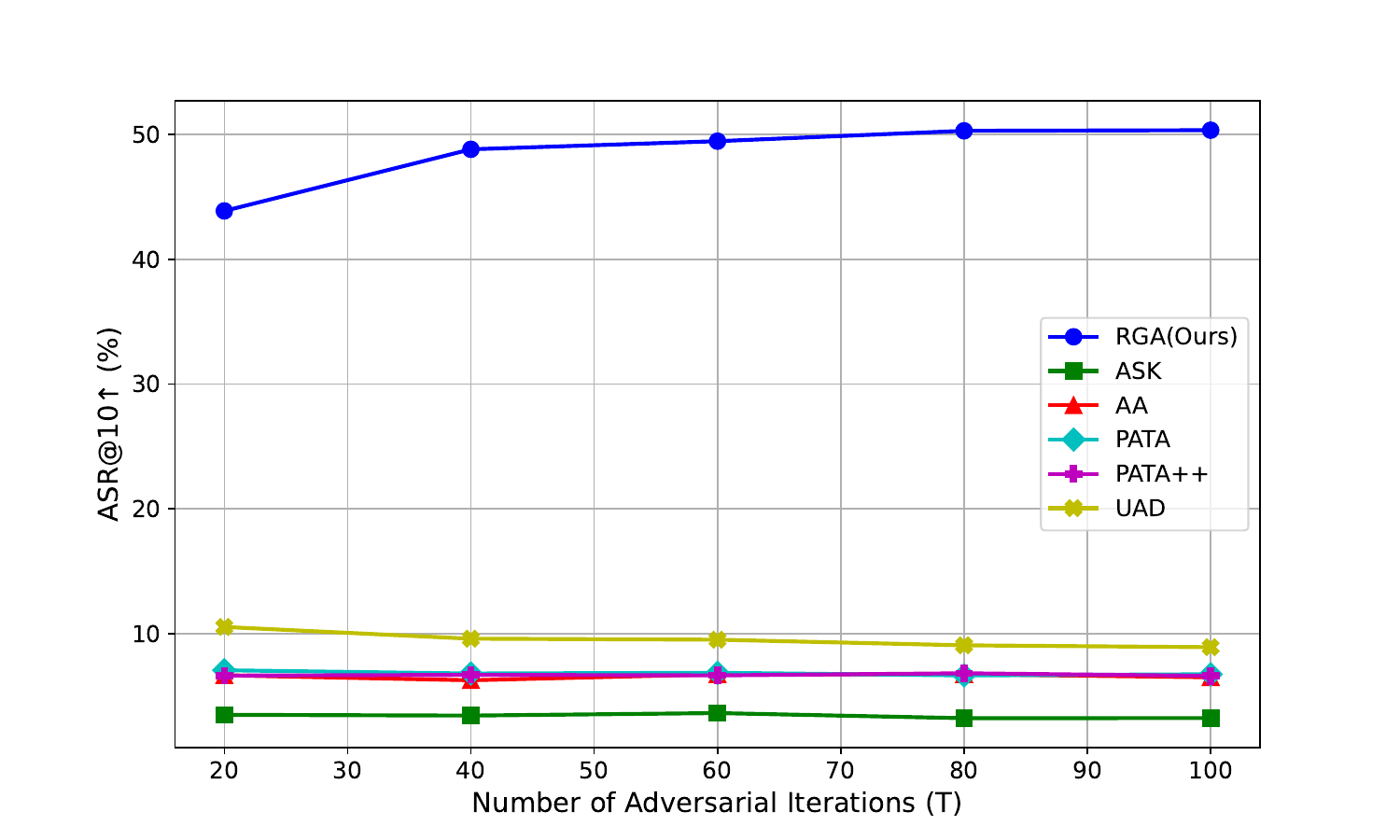}}
    \caption{Sensitivity analysis results of the number of adversarial iterations, $ T $.}
    \label{fig:iterations}
\end{figure*}
\subsubsection{Number of Adversarial Iterations $ T $}
The number of adversarial iterations $ T $ refers to the count of optimization steps taken during the perturbation generation process. As shown in Figure~\ref{fig:iterations}, increasing $ T $ initially improves the attack's success, as it provides more opportunity for refining the adversarial perturbations. However, after a certain point, increasing $ T $ further may lead to overfitting to the specific model, thereby reducing the generalizability of the perturbations. To balance effectiveness and computational cost while avoiding overfitting, we set the default value of $ T $ to 40, as it offers a good compromise between attack success and robustness.

\begin{figure*}[!ht]
    \centering
    \subfigure[mIoU for different values of  $ \gamma $]{
		\includegraphics[width=0.45\linewidth]{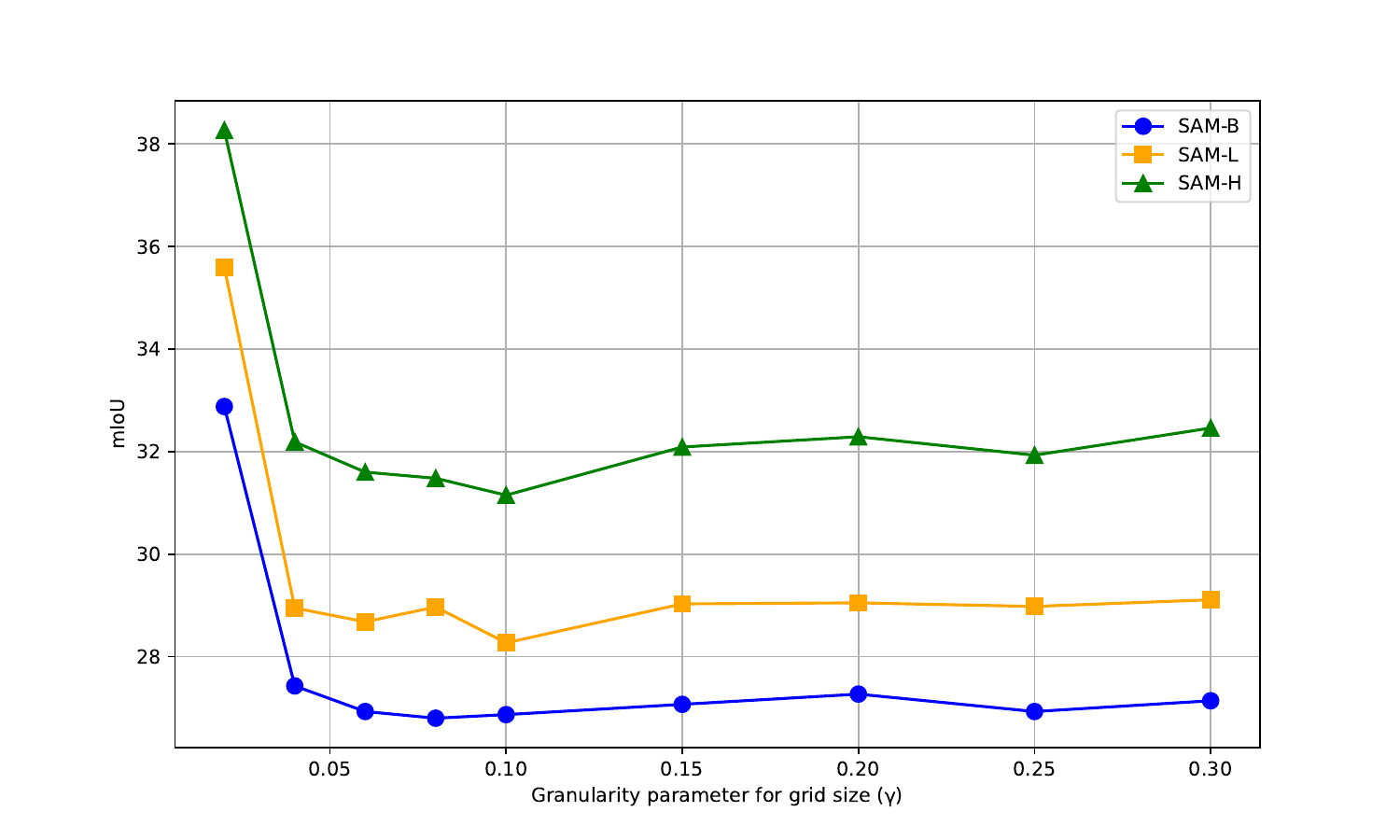}}
    \subfigure[ASR for different values of  $ \gamma $]{
		\includegraphics[width=0.45\linewidth]{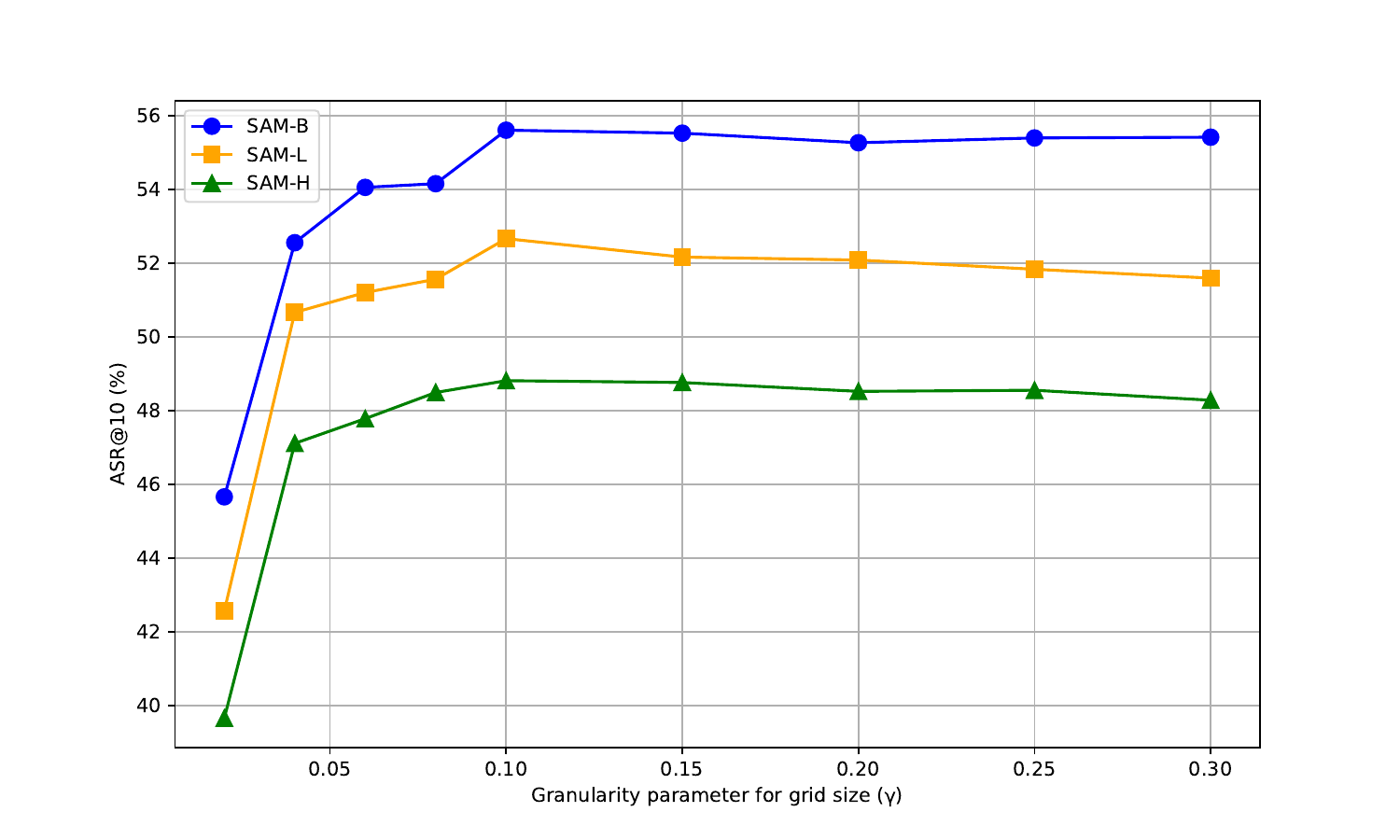}}
    \subfigure[mIoU for different values of  $ n $]{
		\includegraphics[width=0.45\linewidth]{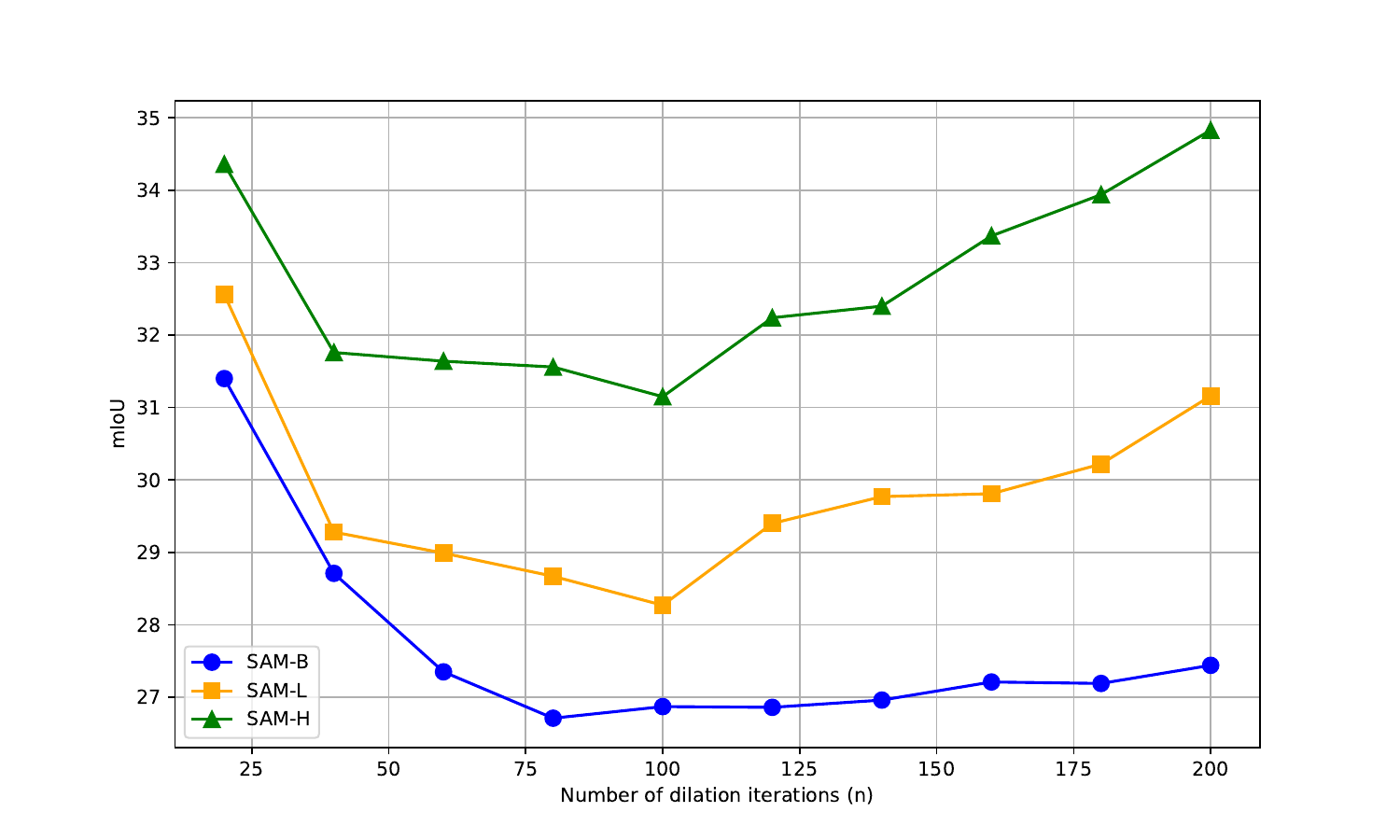}}
  \subfigure[ASR for different values of  $ n $]{
		\includegraphics[width=0.45\linewidth]{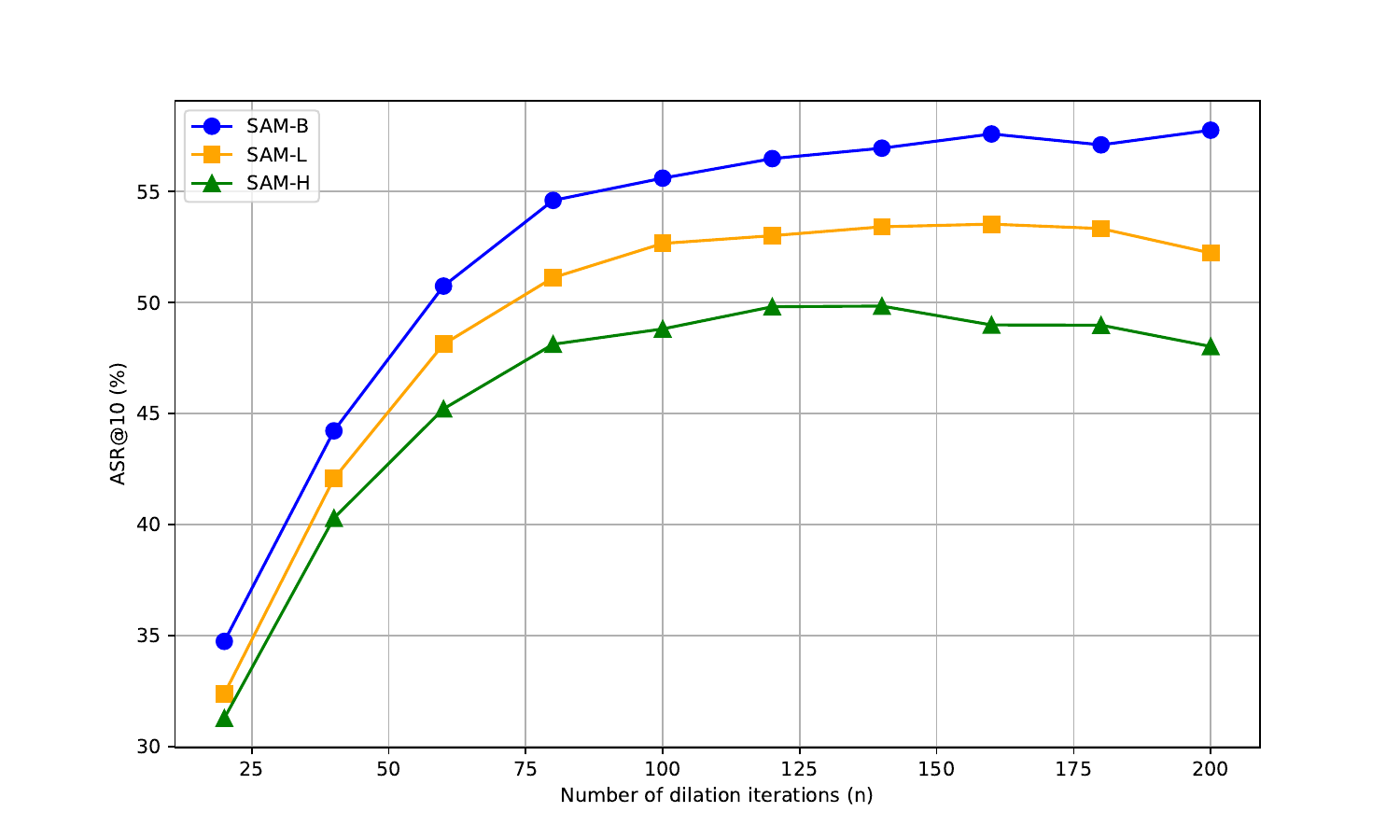}}
    \caption{Sensitivity analysis results of the granularity parameter for grid size, $ \gamma $, and number of dilation iterations $ n $.}
    \label{fig:hy_rga}
\end{figure*}
\subsubsection{Granularity Parameter for Grid Size $ \gamma $ and Number of Dilation Iterations $ n $}
Figure~\ref{fig:hy_rga} presents the sensitivity analysis results for the granularity parameter for grid size $\gamma$ and the number of dilation iterations $n$. The granularity parameter for grid size $ \gamma $ and the number of dilation iterations $ n $ are used to generate the Region-Guided Map (RGM), which plays a crucial role in misleading the SAM model. These parameters determine how the segmentation results are manipulated to create the adversarial effect.

The parameter $ \gamma $ controls the size of the grid blocks used for dividing large segmented regions. If $ \gamma $ is too small, the large regions may be excessively subdivided, which could result in an ineffective adversarial influence that fails to mislead SAM consistently. On the other hand, if $ \gamma $ is too large, the subdivision may be too coarse, reducing the effectiveness of targeting specific parts of the large regions, thereby limiting the ability of RGM to mislead SAM effectively. Hence, finding an appropriate value for $ \gamma $ is key to achieving a balance that maximizes the adversarial impact while maintaining the perceptual quality of the guidance.

Similarly, the number of dilation iterations $ n $ influences how much smaller segmented regions are expanded. If $ n $ is too small, the dilation effect may be insufficient, limiting the impact on the smaller regions that need to be expanded for effective manipulation of the segmentation. Conversely, if $ n $ is too large, the dilation could overly expand these small regions, making the adversarial perturbation too obvious or even causing unintended merging of regions, which reduces the precision of the adversarial attack. Thus, an appropriate value for $ n $ is essential to ensure that the expansion is effective enough to influence the segmentation without introducing unintended side effects.

In summary, the values of $ \gamma $ and $ n $ directly impact the construction of the RGM, which is used to guide SAM to make incorrect segmentations. Both parameters must be tuned carefully to strike a balance between effective misleading of the model and maintaining the subtlety of the perturbations.

\section{Discussion}

The Region-Guided Attack (RGA) employs a powerful approach to adversarial attacks on segmentation models by generating the Region-Guided Map (RGM) through its Segmentation and Dilation (SAD) strategy. While RGA demonstrates considerable strength in degrading segmentation quality by manipulating regions based on size, certain limitations restrict its effectiveness in more specialized segmentation scenarios. Here, we discuss two key limitations: challenges with overlapping regions and the difficulty of achieving desired errors with subtle perturbations.

\textbf{1. Limitations with Overlapping Regions:}
In tasks where regions frequently overlap—such as medical imaging, multi-layered materials in industrial inspections, or semi-transparent objects in computer vision—segmentation often requires intricate, multi-layered outputs that capture the depth and interaction of overlapping structures. RGA’s SAD strategy, which applies binary transformations (dilation for small regions and fragmentation for large ones), is not inherently equipped to manage these complex, overlapping relationships within the segmentation output.

Overlapping regions require an adversarial approach that can distinguish between intersecting areas and selectively perturb them without compromising the multi-layered representation. However, SAD’s current design lacks the granularity to handle overlapping regions effectively. When dealing with complex intersections, SAD’s dilation and fragmentation may lead to coarse distortions that fail to influence the segmented regions in a meaningful way. For example, in a layered tissue sample in medical imaging, disrupting the representation of one tissue layer without affecting the layers above or below requires finer control than SAD’s region-specific transformations currently offer. As a result, RGA’s impact may be limited in these scenarios, as it cannot fully exploit the segmentation model’s sensitivity to overlapping structures.

Future iterations of RGA could address this limitation by incorporating more advanced segmentation-sensitive perturbations that can account for multi-layered and semi-transparent relationships within the data. Approaches like hierarchical perturbation strategies or adaptive transformations that respect the contextual interactions of overlapping layers could improve RGA’s effectiveness in these high-precision tasks.

\textbf{2. Subtle Perturbations May Not Yield Desired Errors:}
RGA’s perturbation approach, while effective for disrupting larger segmented regions or inducing shifts in boundaries, may struggle in cases where minimal yet precise boundary modifications are necessary to degrade the model’s output meaningfully. In domains like high-resolution medical imaging or satellite analysis, segmentation accuracy hinges on the precise identification of fine-grained boundaries, where even small misclassifications can have significant implications. However, the SAD strategy’s binary handling of regions lacks the subtlety needed to create these precise boundary distortions.

SAD’s transformations—dilating small regions and fragmenting large ones—are relatively coarse and may introduce perceptible yet ineffective changes in the segmentation output. When applied to highly detailed regions, these transformations could result in visible but inconsequential alterations that do not meaningfully impact the model’s accuracy. For example, in tasks that require distinguishing between intricate, closely related textures or subtle edge boundaries, RGA’s SAD-based perturbations may produce overly generalized errors that fail to impact model predictions at the desired level of detail.

Addressing this limitation would require an enhancement to SAD’s perturbation methods, allowing it to achieve fine-grained distortions capable of subtly shifting boundaries without creating overtly visible artifacts. Future research could explore integrating more sophisticated techniques such as texture-sensitive perturbations or boundary-preserving modifications to better control the precision of attacks. These enhancements would enable RGA to more effectively target detailed segmentation tasks that demand high fidelity and minimal perceptual distortion.

In summary, while RGA offers a robust, region-guided framework for adversarial attacks, its current SAD strategy presents challenges in tasks involving overlapping regions and in applications where subtle, precise boundary distortions are essential. Future developments in RGA could benefit from adaptive, multi-layered perturbation techniques to handle overlapping regions more effectively and from refined boundary-sensitive transformations to achieve desired segmentation errors with greater subtlety. By addressing these limitations, RGA’s applicability could be extended to a broader range of specialized, high-precision segmentation tasks.

\section{Conclusion}

In this work, we introduced the Region-Guided Attack (RGA), an innovative adversarial attack method designed specifically for segmentation models like the SAM. RGA stands out by generating a Region-Guided Map (RGM) through the Segmentation and Dilation (SAD) strategy, enabling a prompt-agnostic approach that disrupts SAM’s segmentation accuracy. SAD tailors the perturbation strategy according to the size and structure of each segmented region, dilating small regions to exaggerate their presence and fragmenting larger regions to destabilize their boundaries. The resulting RGM acts as a blueprint that guides adversarial perturbations in a spatially targeted manner, enhancing the effectiveness and precision of the attack.

Our quantitative and qualitative evaluations demonstrate that RGA significantly degrades segmentation performance across multiple models and scenarios, achieving high attack success rates in both white-box and black-box settings. By leveraging the feature-based structure of SAM, RGA ensures that perturbations are highly transferable and effective even without prompt-based guidance, highlighting a fundamental vulnerability in segmentation models that rely on spatial coherence.

The success of RGA underscores the need for future research to address region-specific adversarial threats within segmentation frameworks. Defensive measures should focus on enhancing segmentation model robustness against feature-guided perturbations, which are particularly challenging to detect and mitigate. Overall, RGA presents a significant advancement in adversarial attack strategies, providing valuable insights into the structural vulnerabilities of segmentation models and setting a foundation for developing more resilient and secure segmentation algorithms in critical applications.

\section*{Acknowledgments}
This work was supported in part by the STI 2030-Major Projects of China under Grant 2021ZD0201300, and by the National Science Foundation of China under Grant 62276127.

 \bibliographystyle{elsarticle-num} 
 \bibliography{cas-refs}





\end{document}